\documentclass[a4paper]{llncs}
\usepackage{placeins}

\usepackage{amssymb}
\usepackage{url}
\usepackage[pdftex]{graphicx}

\urldef{\mailsa}\path|{Andrew.Lensen, Bing.Xue, Mengjie.Zhang}@ecs.vuw.ac.nz|    
\newcommand{\keywords}[1]{\par\addvspace\baselineskip
\noindent\keywordname\enspace\ignorespaces#1}

\usepackage{color}		%% To change the text colour
\usepackage{algpseudocode, algorithm}
\usepackage{multirow}
\usepackage{amsmath}
\usepackage[caption=true,font=footnotesize]{subfig}

\usepackage[justification=centering]{caption} 
\usepackage{framed}
\usepackage{gensymb}
\usepackage{tabularx,booktabs}
\usepackage{rotating}
\usepackage{float}
\usepackage{cite} %citation orders
\usepackage{afterpage}

\begin{document}

\mainmatter  % start of an individual contribution

%% ========================================
%%		TITLE
%% ========================================
\title{Generating Redundant Features with\\ Unsupervised Multi-Tree Genetic Programming}
%\titlerunning{Generating Redundant Features with Unsupervised Multi-Tree Genetic Programming}
\author{Andrew Lensen\and Bing Xue\and Mengjie Zhang}
%\authorrunning{A.Lensen, B.Xue, and M.Zhang}

\institute{School of Engineering and Computer Science,\\
		Victoria University of Wellington, PO Box 600, Wellington 6140, New Zealand\\
	%}
\mailsa
}
%\author{
%	\author{No Authors}
%	\authorrunning{No Authors}
%	\institute{No Affiliations\\
%		\hfill\\
%		\hfill
%	}
%}

%\toctitle{Lecture Notes in Computer Science}
%\tocauthor{Authors' Instructions}
\maketitle

\begin{abstract}
Recently, feature selection has become an increasingly important area of research due to the surge in high-dimensional datasets in all areas of modern life. A plethora of feature selection algorithms have been proposed, but it is difficult to truly analyse the quality of a given algorithm. Ideally, an algorithm would be evaluated by measuring how well it removes known bad features. Acquiring datasets with such features is inherently difficult, and so a common technique is to add synthetic bad features to an existing dataset. While adding noisy features is an easy task, it is very difficult to automatically add complex, redundant features. This work proposes one of the first approaches to generating redundant features, using a novel genetic programming approach. Initial experiments show that our proposed method can automatically create difficult, redundant features which have the potential to be used for creating high-quality feature selection benchmark datasets.

\end{abstract}

\section{Introduction}
\label{sec:introduction}

Feature Selection (FS) techniques aim to remove features from a dataset which are less useful than others. \cite{liu2012feature} Removing such features can improve the results of the data mining task being performed on the dataset, as well as making the results and/or model produced more interpretable and less complex. Features that should be removed are usually categorised as \textit{irrelevant} or \textit{redundant} features \cite{tang2014feature}.

Irrelevant (or \textit{noisy}) features are those which add little or no meaningful value to a dataset. In the worst case, an irrelevant feature may actually mislead the data mining process, when it contradicts the information given by other ``correct'' features. Removing such features reduces the search space of the data mining task, generally improving performance \cite{liu2012feature}. Redundant features (\textit{r.fs}) share a high amount of information overlap with other features. Removing r.fs can simplify the solutions found (as only one of a set of r.fs is needed), while again reducing the search space of the data mining task \cite{tang2014feature}. In certain cases, results may also be improved by reducing the bias towards a set of very similar features.

Many FS algorithms have been proposed, which are usually evaluated based on how well they can reduce the feature set size, while maintaining (or improving) the results of the data mining task. One technique used to compare FS algorithms is to purposefully add ``bad'' features to a dataset, so that a FS algorithm can be evaluated based on how well it removes those known bad features. Introducing irrelevant features to a dataset is quite straightforward --- choose some stochastic noise generator, and generate a number of noisy features. Introducing r.fs, however, is much trickier, as discussed below.

Perhaps the most naive way of creating a r.f (Y) from a given \textit{source} feature (X) is to multiply each feature value of X by some multiple $\alpha$, such that the $i^{th}$  value of Y is computed as $Y_i = \alpha X_i$. By varying $\alpha$, one can easily generate any given number of r.fs based on $X$. A particularly straightforward method is to simply duplicate features (i.e.\ let $\alpha = 1$) --- but these are trivial to remove. To make the redundancy weaker, one can introduce some bias ($\beta$) such as adding a constant value to each $Y_i$, e.g. $Y_i = \alpha X_i + \beta$. However, such approaches have a number of serious limitations.

The above types of r.fs have very simple redundancies that do not represent realistic interactions between features in real data mining problems. For example, in a dataset of people, two potential features may be an individual's age and income. It is generally true that the older a person, the more they earn, and so we may expect these features to be linearly redundant. However, a child is likely to have no income regardless of their exact age, and a pensioner is likely to have a similar income to others aged over 65. While these two features are certainly partially redundant, the interaction is clearly more complex. In most datasets, the redundancy between two features tend to be even more complex still.
Removing r.fs that have linear redundancies is also quite a trivial FS problem, and so is not an adequate challenge for non-trivial FS algorithms. For example, a greedy algorithm which uses Pearson's correlation can easily find groups of linearly-redundant features by measuring the correlation of each feature to those already selected.

There is hence an obvious need to have methods available to generate r.fs with  (arbitrarily) complex interactions in order to benchmark FS methods more effectively. There has been very little work in the literature that has investigated how to automatically generate non-trivial r.fs. One common method that is used to automatically create functions to perform a particular task is Genetic Programming (GP), an Evolutionary Computation (EC) technique which evolves tree-like functions (\textit{programs}) with a flexible structure. We believe that GP has the potential to evolve functions to produce r.fs, by taking a source feature as the program's input, and producing a r.f as the program's output.

\subsection{Goals}
In this paper, we propose the first approach to automatically generating r.fs, using Genetic Programming for Redundant Feature Creation (GPRFC). The proposed method uses GP to automatically generate functions to produce new r.fs from a given source feature, by using a multi-tree GP representation with a Mutual Information (MI)-based fitness function. This paper will:

\begin{itemize}
	\item Introduce a novel multi-tree GP representation for automatically evolving multiple redundant features from a source feature.
	\item Formulate an appropriate fitness function for evolving high-quality redundant features, using mutual information as a proxy for measuring redundancy.
	\item Provide evidence that the redundant features created are non-trivial and highly redundant.
	\item Analyse a sample of the created redundant features to investigate how their design may introduce redundancy.
\end{itemize}

%\subsection{Organisation}
%To do (optional).

\section{Background}
This section will introduce some core concepts of feature manipulation and mutual information, and briefly discuss some related work.
\subsection{Feature Manipulation}
Feature manipulation is the act of purposefully altering the feature set of a dataset in order to improve the outcomes of a machine learning task. The two most common categories of feature manipulation are feature selection (FS) and construction (FC) \cite{liu2012feature}. FS attempts to select an optimal subset of features in order to improve performance and decrease complexity, whereas FC improves performance by creating new, more powerful high-level features which combine multiple features in some way. 

EC algorithms have seen significant success recently in their application to FS and FC problems, due to their ability to search a large search space effectively \cite{xue2015survey,espejo2010survey}. In particular, Particle Swarm Optimisation (PSO) and Genetic Algorithms (GA) have been widely used for FS, whereas tree-based GP has seen significant use in FC due to its dynamic model structure and ability to apply a variety of functions to the feature set.
\subsection{Mutual Information}
Mutual Information (MI) \cite{jaynes1957information} is an important concept in the field of Information Theory. MI is used as a way to measure the amount of information shared by two variables (or features). In this way, it is a measure of the mutual dependence of two variables, and is one way to measure how redundant one feature is with respect to another --- the higher the MI, the more redundant the features are said to be. MI is formalised as follows:

\begin{equation}
\label{mi}
MI(X,Y) = H(X) + H(Y) - H(X,Y) 
\end{equation}
where the entropy of a feature $X$, $H(X)$, is defined as:
\begin{equation}
H(X) = - \sum_{x\in X} p(x) \times \log_2 p(x)
\end{equation}
and the joint entropy of two features, $X,Y$, is:

\begin{equation}
H(X,Y) = - \sum_{x\in X}\sum_{y \in Y} p(x,y) \times \log_2 p(x,y)
\end{equation}

Equation \ref{mi} can be expanded as follows:

\begin{equation}
MI(X,Y) = - \sum_{x \in X, y \in Y} p(x,y) \times \log_2 \frac{p(x,y)}{p(x)(y)}
\end{equation}

The above definition of MI assumes that the two features have discrete values; in the case of continuous features (such as in this work), the below definition applies:

\begin{equation}
MI(X,Y) = \int_X \int_Y p(x,y) \times \log_2 \frac{p(x,y)}{p(x)(y)} dx\ dy
\end{equation}

Calculating the MI of two continuous features requires knowing the marginal and joint probability density functions (\textit{pdf}) of the two features. In practice, this is infeasible, as the feature values for a given feature can be thought of as only a sample of the underlying \textit{pdf} \cite{kraskov2004estimating}. As such, a number of MI estimators have been proposed for estimating the MI of two continuous features. One venerable method uses a nearest-neighbour estimation approach, which compares the similarity of neighbours for each instance across the two dimensions $X$ and $Y$ to gauge the strength of the relationship between $X$ and $Y$ \cite{kraskov2004estimating}. We use this approach, implemented in the Java Information Dynamics Toolkit (JIDT) \cite{lizier2014jidt}, in this work.
\subsection{Related Work}
As this is the first work to propose the use of an EC algorithm to automatically evolve redundant features, there is no directly related work to discuss. Instead, we will briefly survey the use of GP for FC, since this is the most related area of research to the ideas proposed in this paper.
%, and the use of EC techniques to perform FS, using MI.

A variety of tree-based GP approaches to FC have been proposed, including for problems such as classification and clustering \cite{tran2016genetic,lensen2017gpgc}. Most work uses a representation where a single GP tree produces a single constructed feature, as the output of the tree. The input to the tree is generally the set of features, and an optional random value input. This representation has been extended so that multiple features may be constructed in a single GP individual, commonly using a multi-tree representation \cite{muni2006genetic,lensen2017gpgc}. Other representations have also been proposed \cite{espejo2010survey}, including using multiple sub-trees as a set of constructed features \cite{Ahmed2014Multiple,tran2016genetic}, using specially-tailored node designs \cite{YunZhang2004Multiple}, cooperative co-evolutionary GP \cite{Lin2005Evolutionary}, and even by performing multiple GP runs (each producing a single constructed feature) \cite{neshatian2012filter}.
 These works share similarity with this paper in that they perform a transformation of the original feature space, but they do so in order to improve the performance of a data mining task, rather than to perform feature creation.

\section{The Proposed Method: GPRFC}
\label{sec:the_proposed_method}
This section details the proposed method for automatically generating redundant features, including the GP representation, fitness function, and other important considerations made when designing the method.

\subsection{Genetic Programming Representation}
In this work we use a multi-tree GP representation, where each GP individual contains $n$ distinct trees rather than a single tree. Each tree in an individual represents a single mapping (function) from the source feature ($X$), to a new redundant feature ($Y$). Using a multi-tree representation allows us to generate multiple r.fs per source feature, while encouraging each r.f to be distinct (less redundant) from each other r.f. By generating a variety of r.fs, we increase the diversity of the types of redundancies between the source and redundant features. For example, a r.f $Y_1$ may have a polynomial relationship with $X$, whereas a second r.f $Y_2$ could have an exponential or trigonometric relationship --- both $Y_1$ and $Y_2$ are highly redundant with $X$, but less redundant with each other. This behaviour is encouraged by the fitness function, which will be discussed in more detail in Section \ref{sec:fitnessFunction}.

\subsection{Function and Terminal Sets}
We use only a single terminal in this work: the source feature, $X$. We purposefully do not use a random value input (unlike many GP works), as such a value is unlikely to meaningfully increase MI, and increases the search space unnecessarily. 

In designing the function and terminal sets, it is important to have a wide range of operators with distinct behaviours, so that a variety of redundancy relationships can be constructed in different trees. Based on this, we use a range of different arithmetic, trigonometric, and conditional operators as follows:

\begin{itemize}
	\item Unary operators (taking one input): $\sin(a)$, $\tan(a)$, $\tanh(a)$, $\log(a)$, $e^a$, $\sqrt{a}$, $a^2$, $a^3$, $-a$. We purposefully exclude $cos(a)$ due to its similarity to $sin(a)$. While $a^2$ and $a^3$ can be easily constructed in a GP tree, we include them as useful ``building blocks''.
	\item Binary operators (with two inputs): $a+b$, $a\times b$, $\max(a,b)$, $\min(a,b)$,  $a^b$. We exclude $a-b$ and $a \div b$ as they are the complements of addition and multiplication, and as they were found to negatively affect the learning process by easily producing constant values (i.e.\ $X-X = 0$, $X\div X = 1$).
	\item A single ternary operator, $if$, which outputs the second input if the first input is non-negative and the third input otherwise. This operator, in addition to $\max$ and $\min$, allows complex conditional behaviour and non-continuous functions to be generated.
\end{itemize}

%%scale 
%The fitness function in this work is designed to encourage a \textbf{high} level of redundancy between the source and generated features.

%, while encouraging a \textbf{low} amount of redundancy between different generated features.

\subsection{Fitness Function}
\label{sec:fitnessFunction}
Our proposed fitness function is based on the concept of Mutual Information, a measure of the dependency between two features. We use MI as a proxy to measure the redundancy of of a generated feature: if the MI between the source and generated feature is high, the generated feature is said to be highly redundant. Hence, the MI between the source and each generated feature/tree should be \textbf{maximised}. In addition, we choose to \textbf{minimise} the MI between each pair of generated features. In doing so, we implicitly encourage a set of r.fs that are redundant in \textit{different ways} to be generated --- for example, if two r.fs both had linear redundancies with the source feature, they would also have a high MI between them. This decision automatically increases the complexity of the generated r.fs, which should also make them harder for FS algorithms to remove. We describe the formulation of the fitness function in detail below.

Let $X$ be the source feature, $I$ be the GP individual whose fitness is being measured, which contains a set of trees ($T$), where $n$ is the number of trees. Let the ``baseline'' MI, $\Psi$ (used as a normalisation factor), be defined as the output of the MI estimation algorithm for $\Psi = {MI}(X,X)$. In measuring the quality of $I$, we consider the \textbf{minimum} MI between any $X$ and any r.f (called minSourceMI), as well as the \textbf{maximum} mean MI between any r.f and all other r.fs (called maxSharedMI). The quality of $I$ is measured by how much more redundant the r.fs are with $X$ than with each other, defined as follows:

\begin{equation}
\text{minSourceMI} = \min_{t \in T}\frac{ MI(X,t)}{\Psi}
\end{equation}
\begin{equation}
\text{maxSharedMI} = \max_{t \in T} \frac{\sum_{y \in T, y \neq t}\frac{ MI(t,y)}{\Psi}}{n-1}
\end{equation}
\begin{equation}
\text{Quality}_I = \text{minSourceMI} - \text{maxSharedMI}
\end{equation}

While this quality measure is expected to be suitable as a fitness function, it does not consider that having a minSourceMI below a certain threshold means that the r.fs produced are in fact not very redundant at all. In addition, generally a lower minSourceMI leads to a higher potential fitness, making the fitness function biased towards creating a set of r.fs which are very unrelated to each other, and only weakly related to the source feature. To remedy this, we introduce an additional component to the fitness function for when the minSourceMI is below some threshold, $\Theta$, where $\Theta$ is the minimum ``acceptable'' redundancy between a r.f and $X$. In other words, individuals not meeting this criteria can be thought of as \textit{infeasible solutions}. For these infeasible solutions, we do not consider the shared MI between r.fs to be important, as at least one of the r.fs is not acceptable. To encourage increasing the redundancy of each r.f in this scenario (i.e.\ encouraging the solution towards becoming feasible), we penalise individuals based on the mean MI between the source and each r.f:

%\footnote{In other words, the lower the MI shared with the source feature, the easier it is to create a set of r.fs that share little information themselves. Consider the extreme case where the set of r.fs can be independent (0 MI) at a low enough minSourceMI.}

\begin{equation}
\text{Penalty}_I = \frac{-1}{\text{meanSourceMI}}
\end{equation}
\begin{equation}
\text{meanSourceMI} = \frac{\sum_{t \in T} \frac{MI(X,t)}{\Psi}}{n}
\end{equation}

This penalty function is designed as such so that the higher the meanSourceMI, the lower the penalty applied. Our fitness function is then the combination of these two functions:

\begin{equation}
\text{Fitness}_I = \begin{cases}
\text{Quality}_I,& \text{if } \text{minSourceMI}\geq \Theta\\
\text{Penalty}_I,              & \text{otherwise}
\end{cases}
\end{equation}

As the Penalty term of the fitness function is constrained to be less than 0, an individual with $\text{minSourceMI}\geq \Theta$ will nearly always be better than one that does not meet the $\Theta$ threshold. As our measurements of MI are normalised by $\Psi$, the threshold $\Theta$ can be chosen (roughly) from the range $[0,1]$, where a value of $\Theta = 0$ corresponds to all r.fs being independent to $X$, and a value of $\Theta = 1$ corresponding to all r.fs being perfectly redundant with $X$. In practice, we found a $\Theta$ in the range $[0.6,0.7]$ was a good choice for $n=5$.

%Note that as the MI estimator is an approximation, and as MI is not a linear relationship\footnote{I don't believe that 80\% MI means strictly twice as much redundancy as 40\% MI...which does bring the fitness function into question?}, the meaning of a given $\Theta$ should not be over-analysed.
\subsection{Further Considerations}
A number of other factors had to be addressed in order to achieve good results with the proposed method. These are discussed in turn below.
%This section discusses a number of other decisions made and techniques found to overcome a number of limitations of the proposed approach. They are as follows:

%\begin{itemize}
	%\item
	 To improve the consistency of the GP method, the source feature was scaled so that all values fall in the range $[0,1]$. However, this meant that at least one source feature value would be exactly $0$. An input of $0$ to the GP tree was found to significantly affect training as it would often result in multiplication or division by $0$ within the tree. The common occurrence of dividing by $0$ was particularly troublesome, as it meant the tree would not produce a valid output, making the whole individual invalid. To remedy this, we added a small weighting to each feature value, of size $\epsilon$, such that all feature values lie in $[0+\epsilon,1+\epsilon]$. In this work, we setting found $\epsilon = 1\times10^{-3}$ to be suitable.
	%\item 
	
	While the above scaling approach is expected to work well on artificially-generated datasets, it does not address an issue with many real-world classification datasets: duplicate feature values. Consider the example of a (real-world) dataset where a feature takes values in $\{1,2,3,4\}$. Given there are only 4 unique inputs to a GP tree, the tree may only produce (at most) 4 unique outputs. This greatly limits the ability of GP to learn to create multiple distinct r.fs as only very ``coarse'' r.fs can be generated (with low complexity). To address this performance limitation, we add a small amount of stochastic noise (using a constant seed) to each source feature value, so that each feature value is likely to be distinct. This is essentially equivalent to changing the input of the GP tree to be $X + \delta$, where $\delta$ is a small value which is consistent for a given value of $X$. As before, we ensure $\delta$ is strictly positive. The feature values are hence in the range $[0 + \delta, 1+\delta]$, where we defined $\delta$ to be a random number between $0.001\epsilon$ and $\epsilon$. In both the above approaches, we still evaluate the MI between a r.f and $X$ (i.e.\ when computing the fitness function) using the \textbf{original} (i.e.\ unscaled) feature values, to ensure we measure the true redundancy.
	
	%\item 
	In addition to scaling the source feature, we also scale the constructed redundant features to lie in $[0,1]$. This serves two purposes: it ensures the r.fs have ``sensible'' ranges, and so can be more easily visualised, and it also means they have the same range as the source feature, which is important for many algorithms such as $k$-nearest neighbour, $k$-means clustering etc. Finally, the redundant features are rounded to $5$ decimal places, to prevent GP from evolving very sensitive features whose precision may be lost when saved to file or used in another algorithm.
	
	\subsubsection{Other Parameter Settings:}
	 We use a relatively high max tree depth of 15 and mutation rate of 40\% (with crossover of 60\%). Using a high max tree depth was found to encourage more complex trees to be formed, which tended to produce more complex features. Evaluating the larger trees is not significantly more costly, as the computation of MI is the most expensive part of the fitness evaluation. 40\% mutation was used to encourage the generation of more diverse trees --- however, crossover is still important to ensure that useful function ``building blocks'' are passed between different GP individuals. The population size was set to 1,024, and top-10 elitism was used, as standard. In this work, we used $n=5$ trees as it was found to produce a reasonable balance between making a large number of r.fs and making highly diverse r.fs. Decreasing $n$ will produce r.fs which are less redundant to each other, whereas increasing $n$ will give more, but less distinct r.fs. $\Theta$ was set to 0.7 in this work based on empirical results.

\section{Experiment Design}
\label{sec:experiment_design}
We tested the proposed GPRFC approach on a number of popular datasets, as listed in Table \ref{table:datasets}. These datasets include three classification datasets from the UCI repository \cite{uci}, two of which are quite simple and easy to classify well (Iris and Wine), whereas the third (Vehicle) is more challenging. We also use two synthetic clustering datasets (10d10cE and 10d40cE), which have 10 and 40 clusters respectively and are generated using an Ellipsoidal cluster generator \cite{handl2007evolutionary}. The datasets chosen all have a reasonably small number of features to reduce the number of GP runs required. For each dataset, 5 r.fs are created per source feature, to give a result of $d + 5d = 6d$ features for $d$ source features. As the feature creation approach uses GP, it is stochastic, and so at least 30 runs were performed on each dataset.

To evaluate the created r.fs, we used the classifiers, clusterers, and feature selection algorithms provided by the WEKA \cite{hall2009weka} package. We selected four varied and popular classifiers: the J48 Decision Tree (DT) algorithm, $k$-nearest neighbour (KNN, with $k=3$), Naive Bayes (NB), and the Sequential Minimal Optimisation implementation of the Support Vector Machine (SVM). For clustering, we use 3 different varieties of clustering algorithms: $k$-means++, agglomerative clustering (the average-link variant), and the Expectation Maximisation (EM) algorithm. 

\begin{table}[tb]

%	\renewcommand{\arraystretch}{1.1}
%	\fontsize{8}{9}\selectfont
%	\vspace{-2em}
	\caption{Datasets used in the experiments.}
	\label{table:datasets}
	\centering
	%\vn{0.3}
	
	\begin{tabular*}{.7\textwidth}{@{\extracolsep{\fill}}lrrr}
		\toprule
		Name & No. Features & No. Instances & No. Classes/Clusters \\ 
		\midrule
		Iris & 4 & 150 & 3\\
		Wine & 13 & 178 & 3\\
		Vehicle & 18 & 846 & 4\\
		10d10cE & 10 & 2903 & 10\\
		10d40cE & 10 & 2023 & 40\\
		\bottomrule
	\end{tabular*}
	
	\vspace{-1em}
	%\vspace{-1em}
\end{table}

%\subsection{Datasets}
%\label{datasets}

\section{Results and Discussion}
\label{sec:results_and_discussion}

As there are no known redundant feature creation methods which use a guided search to automatically find good r.fs, we are unable to directly compare GPRFC to a known baseline. Instead, we directly evaluate the quality of the r.fs created across the datasets in terms of the fitness achieved. 
We also investigate how the addition of the r.fs affects the performance of some common classification and clustering algorithms, and how well some simple feature selection algorithms are able to identify (and remove) the added r.fs, in order to evaluate the suitability of the proposed method for creating benchmark datasets. 

\subsection{Fitness}
Table \ref{table:allResults} shows the performance of GPRFC in terms of the average fitness achieved across the tested datasets. GPRFC achieves a high fitness on two of the three classification datasets: Iris and Vehicle. A mean fitness of 0.351 on Vehicle indicates that the typical created r.f is 35.1\% more redundant with the source feature than the other created r.fs, for example, 75.1\% MI with the source feature vs only 40\% MI with the other created r.fs. The performance on the two synthetic clustering datasets is not as strong, but the created r.fs are still clearly more redundant with the source feature than each other. 

\begin{table}[tb]
	\caption{Fitness achieved by GPRFC across all features on each dataset. Standard deviation is taken across the means for each feature. At least 30 runs were performed per feature per dataset.}
	\label{table:allResults}
	\centering
	\begin{tabular*}{0.45\textwidth}{@{\extracolsep{\fill}}lrr}
		\toprule
		Dataset & Mean & Std. Dev \\ 
		\midrule
		Iris & 0.333 & 0.082\\
		Wine & 0.203 & 0.055\\ 
		Vehicle & 0.351 & 0.041\\ 
		10d10c & 0.106 & 0.010\\ 
		10d40c & 0.141 & 0.006\\
		\bottomrule
	\end{tabular*} 
\vspace{-1em}
\end{table}

In general, it appears that datasets containing fewer instances tend to have a higher standard deviation --- perhaps as the fitness is more sensitive to any one single feature value being altered during the evolutionary process. The fitness across the Iris dataset, which has the highest standard deviation, is shown in Table \ref{table:irisResults} for each feature. This table clearly shows that F2 has a much lower mean fitness than the other features, and so gives a high standard deviation on the Iris dataset. It is not obvious as to why GPRFC can learn more effectively on certain features. One explanation may be that as GPRFC produces functions that transform the feature space, features that have very dense feature value distributions are harder to transform with a high level of granularity, and so harder to optimise. However, further investigation is needed.

\begin{table}[tb]
	\caption{Fitness achieved by GPRFC across 30 runs on the Iris dataset.}
	\label{table:irisResults}
	\centering
	\begin{tabular*}{0.45\textwidth}{@{\extracolsep{\fill}}lrr}
		\toprule
		Feature & Mean & Std. Dev \\ 
		\midrule
		F0 & 0.362 & 0.050 \\ 
		F1 & 0.398 & 0.036 \\ 
		F2 & 0.213 & 0.029 \\ 
		F3 & 0.359 & 0.053 \\ 
		\bottomrule
	\end{tabular*} 
\vspace{-1em}
\end{table}

\subsection{Classification Performance}
The performance of a number of classifiers on the original datasets compared to the datasets with added r.fs (``augmented datasets'') are shown in Table \ref{table:classificationPerformance}. In general, performance is very consistent between the original and augmented datasets  --- in most cases, dropping by 2-3\%, or holding steady. Given that redundant features aren't inherently misleading to a classifier, it makes sense that performance may not drop much -- though the classification model produced will certainly be more complex. Two major exceptions to this are on the KNN classifier, which had a decrease of around 5\% and 11\% accuracy on Iris and Vehicle respectively. This is likely due to the created r.fs not having the same distances between instances' feature values as the source features had. As KNN is a distance-based classifier, any addition of features which transform the feature space non-linearly will directly alter the distances between instances. Testing on more difficult or datasets with many more features may show a bigger decrease in performance, as the search space may become complex/large enough to better challenge classification algorithms. The small increase in performance on the Vehicle dataset with NB is not statistically significant.

\begin{table}[tb]
	\setlength{\tabcolsep}{3pt}
	\centering
	\caption{Test classification accuracy on each of the datasets before (``Original'') and after (``Augmented'') the created r.fs were added. Each of the 30 runs of GPRFC produced one augmented dataset --- hence, the mean and standard deviation accuracy on these 30 augmented datasets are reported. A split of 70\% training to 30\% test was used.}
	\label{table:classificationPerformance}
	\begin{tabularx}{\textwidth}{lllXllXll}
		\toprule
		\multicolumn{1}{c}{Method} & \multicolumn{2}{c}{Iris} && \multicolumn{2}{c}{Wine} && \multicolumn{2}{c}{Vehicle} \\ \midrule
		& \multicolumn{1}{c}{Original} & \multicolumn{1}{c}{Augmented} && \multicolumn{1}{c}{Original} & \multicolumn{1}{c}{Augmented} && \multicolumn{1}{c}{Original} & \multicolumn{1}{c}{Augmented} \\ \cmidrule{2-3} \cmidrule{5-6} \cmidrule{8-9}
		DT & 0.978 & 0.956$\pm$0.004 && 0.981 & 0.977$\pm$0.017 && 0.709 & 0.692$\pm$0.025 \\
		KNN & 1.000 & 0.947$\pm$0.035 && 0.962 & 0.961$\pm$0.028 && 0.720 & 0.613$\pm$0.029 \\
		NB & 0.978 & 0.964$\pm$0.018 && 1.000 & 0.979$\pm$0.018 && 0.465 & 0.490$\pm$0.025 \\
		SVM & 0.978 & 0.968$\pm$0.020 && 0.981 & 0.974$\pm$0.016 && 0.740 & 0.715$\pm$0.018 \\ \bottomrule
	\end{tabularx}
\vspace{-1em}
\end{table}

\subsection{Clustering Performance}
The performance of three clustering algorithms on the original and augmented datasets was investigated, with the results shown in Table \ref{table:clusteringPerformance}. As with the classification datasets, there is generally little change in performance --- in fact, performance appears to slightly increase when adding the created r.fs. However, the clusters produced are more complex and less interpretable --- with $6d$ features per instance compared to only $d$ in the original datasets.
\begin{table}[tb]
	\setlength{\tabcolsep}{3pt}
	\centering
	\caption{Adjusted Rand Index of the clusters produced on each of the datasets before (``Original'') and after (``Augmented'') the created r.fs were added. Each of the 30 runs of GPRFC produced one augmented dataset --- hence, the mean and standard deviation accuracy on these 30 augmented datasets are reported. $k$-means++ and EM are stochastic algorithms and so the mean of 30 runs per augmented dataset was used.}
	\label{table:clusteringPerformance}
	\begin{tabularx}{0.77\textwidth}{lrrXrr}
		\toprule
		\multicolumn{1}{c}{Method} & \multicolumn{2}{c}{10d10cE} && \multicolumn{2}{c}{10d40cE}  \\ \midrule
		& \multicolumn{1}{c}{Original} & \multicolumn{1}{c}{Augmented} && \multicolumn{1}{c}{Original} & \multicolumn{1}{c}{Augmented} \\ \cmidrule{2-3} \cmidrule{5-6}
		$k$-means++ & 0.548 & 0.558$\pm$0.023 && 0.445 & 0.491$\pm$0.019 \\
		Agglomerative & 0.495 & 0.528$\pm$0.064  && 0.276 & 0.309$\pm$0.046  \\
		EM & 0.588 & 0.606$\pm$0.014 && 0.433 & 0.520$\pm$0.011  \\
		\bottomrule
	\end{tabularx}
\end{table}

%\subsection{Quality of Created Features}
%\subsubsection{In Supervised Learning (Classification)}
%\subsubsection{In Unsupervised Learning (Clustering)}
%\definecolor{darkred}{rgb}{0.55, 0.0, 0.0}
%{\color{darkred}
%\subsection{Feature Selection Results}
% So, I'm not entirely sure what we should do here. A few options:
% 
% \begin{itemize}
% 	\item Analyse a single example of a single dataset in terms of the feature ranking results (e.g.\ using information gain), and feature selection results (e.g.\ using correlation-based FS with a backtracking search).
% 	\item Try to do a more substantial analysis of the various classification datasets. One possibility is performing FS on the original datasets to get the set of ``correct'' selected features. From this, FS could be then be performed on the 30 augmented datasets, and the results could be analysed by seeing how well the FS algorithm selects a single r.f from each r.f ``group''. In other words, if F5 is selected on the original dataset, then only \textbf{one} of $\{F5,F5a,F5b,F5c,F5d,F5e\}$ should be selected in theory. This has the issue however that selecting multiple redundant features could be good for performance (?) so the results may be hard to argue.
% 	\item Leave out this section and position it as future work. This has the issue of making it harder to justify that our created r.fs are actually difficult for FS algorithms to find.
% \end{itemize}
%
% }

\subsection{Feature Selection Results}

\subsubsection{Feature Ranking:}
A common technique used in supervised feature selection is to measure how well a given feature can be used to predict the class label for a set of instances. Information Gain (IG) \cite{jaynes1957information} is often used as a metric to measure this, using similar principles to MI. To see how ``confusing'' our created r.fs may to be a FS algorithm, we ranked the features of the median and best result of applying GPRFC to the Iris dataset, using IG as shown in Table \ref{igTests}. We use the Iris dataset as our example as it has the fewest features, and so can be analysed most easily. The majority of created r.fs have similar rankings to their source features, with the top half of the ranks taken by F2 and F3, and the bottom half by F0 and F1. This is unsurprising --- given that the created r.fs share a high amount of information with the source features, they are likely to also have a similar ability to predict the class label. However, the r.fs do have small variances in their IG value compared to their source features: for example, on the median result, F2 has an IG of 1.418, and its r.fs have IG values between 0.864 and 1.367. On the best result, F2c and F2e actually have \textbf{better} IG than the source feature; F2's r.fs range in IG value from 0.827 to 1.456. These results indicate that while the created r.fs clearly share information with their source features, they are still different enough that their redundancy is non-trivial to identify and they are likely to have an effect on the classification task. 

\begin{table}[tbh]
	\caption{Features ranked by Information Gain (with respect to the class label) on the augmented datasets created by the median (\ref{medianIG}) and best (\ref{bestIG}) runs of GPRFC.}
	\label{igTests}
	\centering
	\subfloat[Median]{
		%	\caption{My caption}
		%\label{my-label}
		\begin{tabularx}{0.25\textwidth}{@{\extracolsep{\fill}}ll}
			\toprule
			Info Gain & Feature \\ \midrule
			1.418     & F2      \\
			1.385     & F3a     \\
			1.378     & F3      \\
			1.367     & F2b     \\
			1.274     & F3c     \\
			1.216     & F2a     \\
			1.163     & F3e     \\
			0.976     & F3d     \\
			0.918     & F2d     \\
			0.908     & F3b     \\
			0.864     & F2e     \\
			0.705     & F0a     \\
			0.698     & F0      \\
			0.554     & F2c     \\
			0.376     & F1e     \\
			0.376     & F1b     \\
			0.376     & F1      \\
			0.364     & F0b     \\
			0.325     & F1d     \\
			0.177     & F0c     \\
			0.118     & F0d     \\
			0.098     & F0e     \\
			0.000     & F1a     \\
			0.000     & F1c     \\ \bottomrule
		\end{tabularx}
	\label{medianIG}
	}\hfil	
	\subfloat[Best]{
	%\caption{My caption}
	\begin{tabularx}{0.25\textwidth}{@{\extracolsep{\fill}}ll}
		\toprule
		Info Gain & Feature \\ \midrule
		1.456     & F2c     \\
		1.421     & F2e     \\
		1.418     & F2      \\
		1.378     & F3      \\
		1.313     & F3a     \\
		1.295     & F2b     \\
		1.214     & F3d     \\
		1.098     & F3c     \\
		1.077     & F3e     \\
		1.057     & F3b     \\
		0.918     & F2d     \\
		0.827     & F2a     \\
		0.722     & F0a     \\
		0.698     & F0      \\
		0.597     & F0e     \\
		0.597     & F0d     \\
		0.419     & F0c     \\
		0.376     & F1c     \\
		0.376     & F1      \\
		0.198     & F1d     \\
		0.158     & F1b     \\
		0.089     & F1a     \\
		0.000     & F0b     \\
		0.000     & F1e     \\ \bottomrule
	\end{tabularx}
\label{bestIG}
}
\vspace{-2em}
\end{table}

\subsubsection{Using a FS algorithm:}
To further investigate how suitable the created r.fs are for benchmarking FS algorithms, we applied a basic FS algorithm to the same two augmented Iris datasets. We used the canonical Sequential Floating Forward Search (SFFS) \cite{pudil1994floating}, which is an extension to the Sequential Forward Search (SFS) algorithm. SFS starts with no features selected, and iteratively adds the best of the remaining unselected features, until performance is not improved by adding the next feature. SFFS follows the same procedure, but also performs a backwards search after each addition of a new feature. That is, it repetitively removes the worst feature in the selected subset, until performance is not improved by removing an additional feature. This floating search helps to avoid the FS algorithm from getting stuck in local optima, and makes SFFS one of the most commonly used deterministic FS algorithms.

In this work, we used a wrapper method where the SVM algorithm is used to classify the dataset for a given feature subset, and the accuracy of the results is used as the performance of the selected features. We use the SVM classifier as it had the highest performance in Table \ref{table:classificationPerformance}. Our SFFS implementation used a training set to train the SVM, and a validation set to test the performance of the SVM on unseen data. The performance of the validation set is the performance of the selected features during training. Finally, we use a separate unseen test set to measure the quality of the final selected features on unseen data. The training, validation, and test sets are 60\%, 20\% and 20\% of the shuffled dataset respectively.
 
On the median dataset (for Iris), this FS method selects features [F2b,F3] with a test accuracy rate of $0.933$. On the best dataset, [F0,F1a,F3] are selected with an accuracy of $0.967$. On the original dataset, the FS method selects only F3, with an accuracy of $0.967$. While the obtained classification accuracy on the augmented datasets is similar, the FS method clearly selects extraneous features, which gives a more complex model than that of when only a single feature is selected on the original dataset.

Given that obtaining good performance on Iris is easy, and so FS is also relatively easy, we performed a similar experiment on the Vehicle dataset to see if different behaviour occurs on a harder, higher-dimensional problem. On the original 18-dimensional vehicle dataset, the SFFS method (as described above) selects  12 features: [F0,F2,F5,F7,F8,F9,F10,F12,F13,F16,F17], with a test accuracy of $0.710$. On the median augmented dataset however, it selects 7 features: [F3,F9b,F12d,F12e,F13,F17,F17e] with a test accuracy of only \textbf{$0.473$}. The FS method has clearly struggled to find a good set of features, as it selects multiple redundant features while also failing to select many features that were selected in the original dataset. Furthermore, the training accuracy was reasonable similar for both datasets ($0.769$ and $0.686$ for the original and augmented respectively), indicating that the created r.fs were able to mislead the FS algorithm well enough to prevent a well-generalised classifier from being produced. Further investigation is needed to provide more quantitative evidence that GPRFC produces r.fs that make difficult benchmark datasets, but the preliminary results are a promising sign that the proposed method has potential.

\section{Further Analysis}
\label{sec:furtherAnalysis}
While we have shown that GPRFC is able to automatically produce a set of redundant features that have high MI with an original feature, it is not yet obvious \textbf{how} it is able to do so. To investigate this aspect, we plotted the created features against the original features for each of the 4 features on the Iris dataset, using the median result of the 30 runs of GPRFC. We choose to analyse Iris as it is the dataset with the smallest feature set. These plots are shown in Figure \ref{irisPlots}.

\def \scaleFactor{0.2}
\begin{sidewaysfigure}[p]
%	\vspace{-1em}
	%\vspace{-10em}
	\centering
	\subfloat{
		\label{f0a}
		\includegraphics[width=\scaleFactor\textwidth]{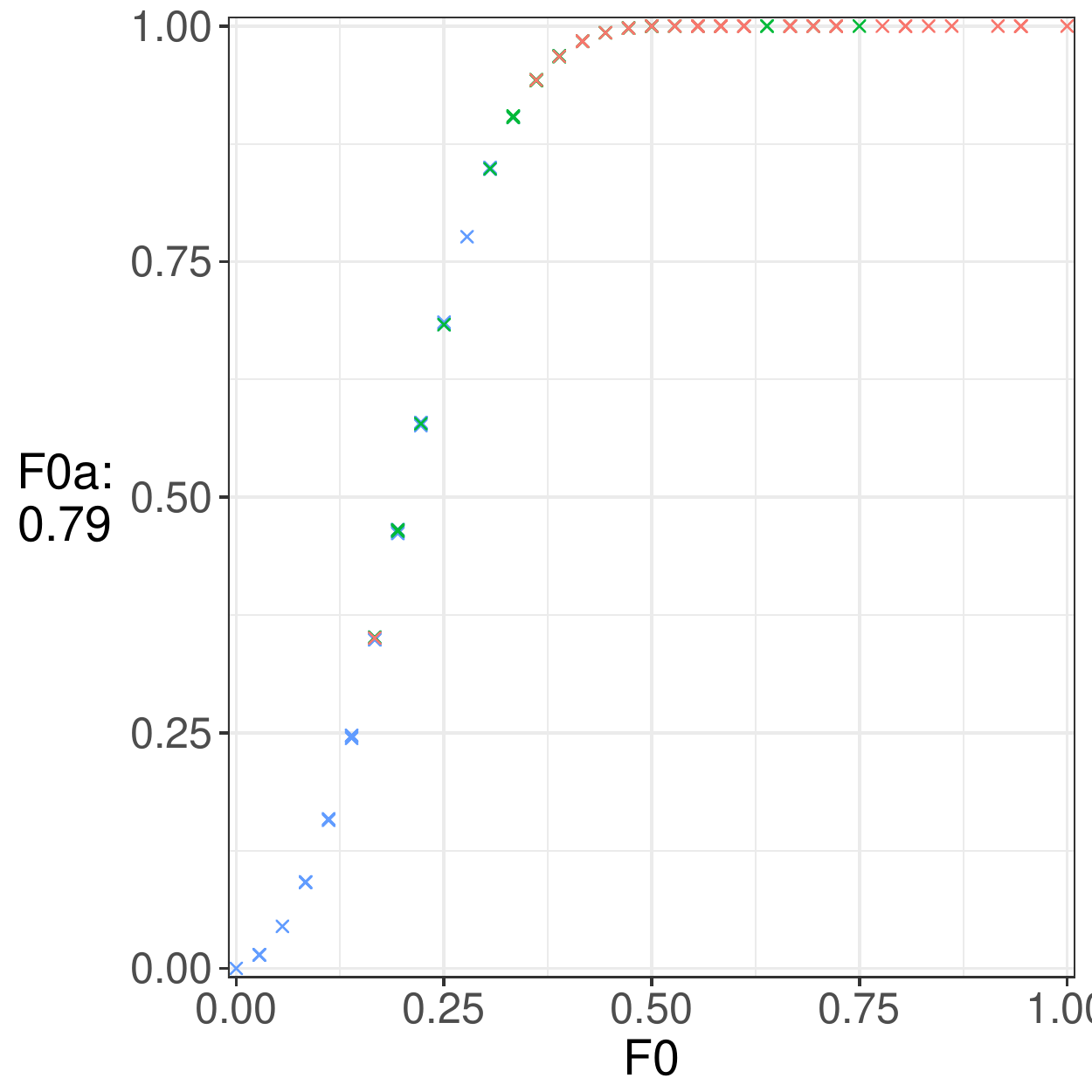}
\hspace{-.25cm}	
} 
	\subfloat{
		\label{f0b}
	\includegraphics[width=\scaleFactor\textwidth]{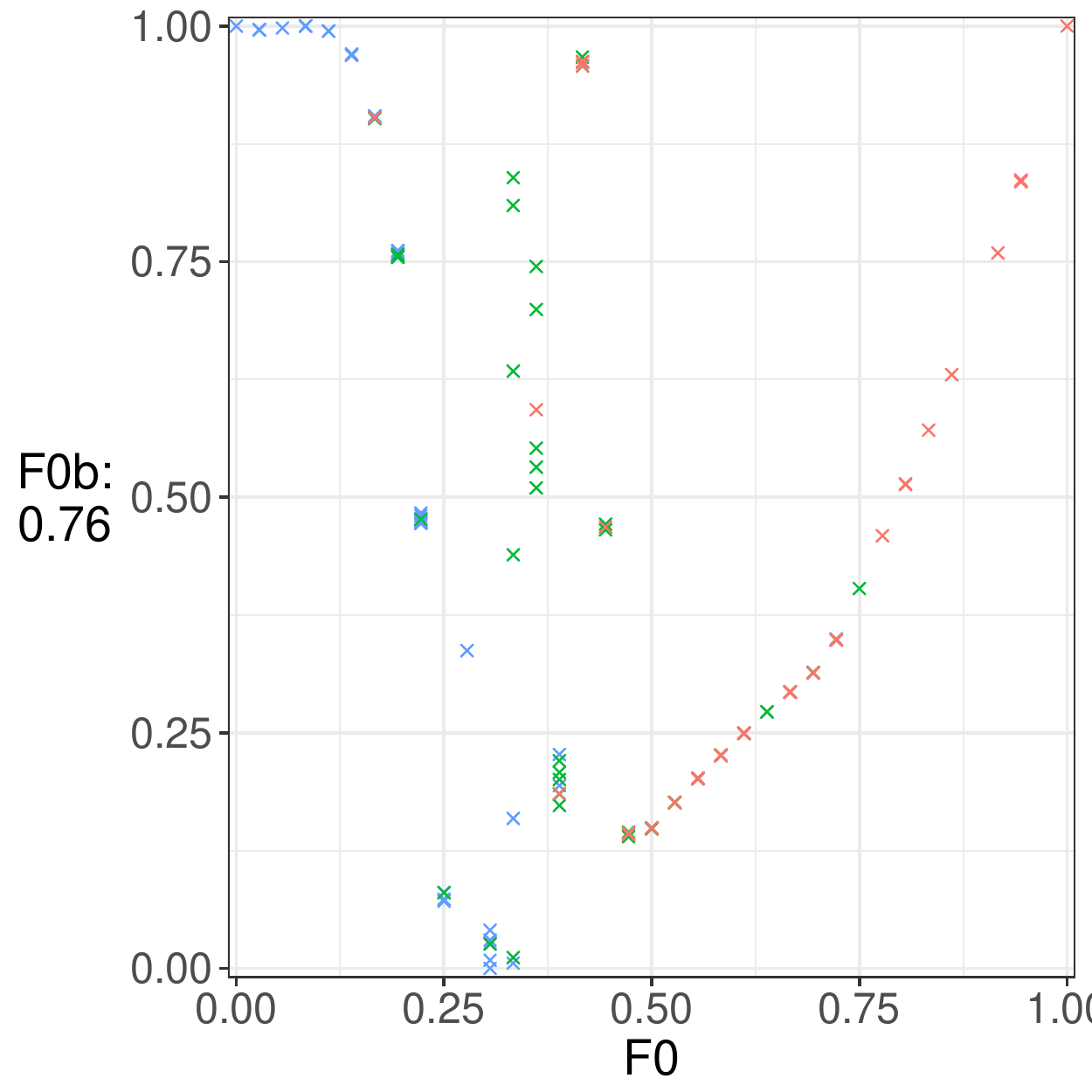}
\hspace{-.25cm}	
} 
\subfloat{
	\label{f0c}
	\includegraphics[width=\scaleFactor\textwidth]{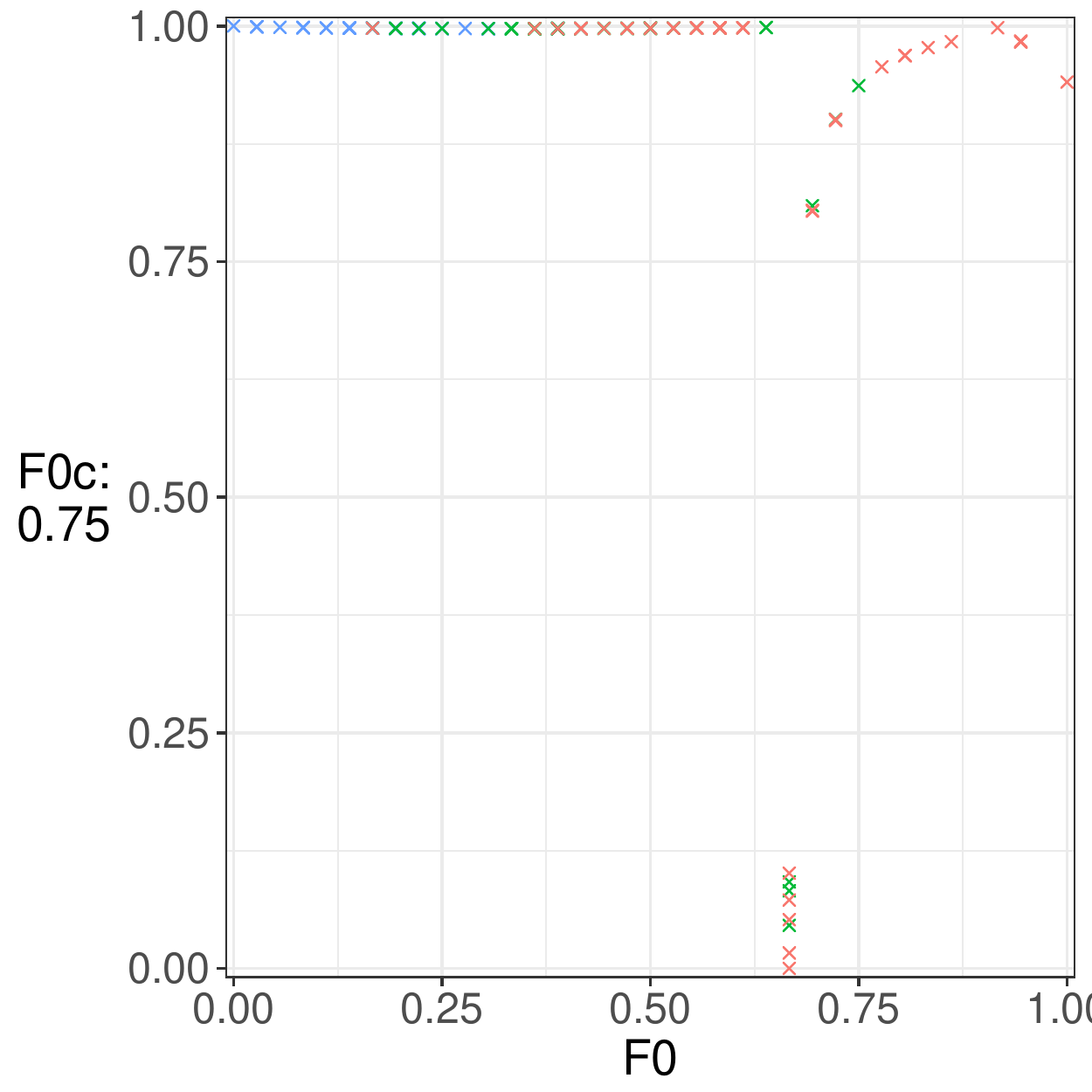}
\hspace{-.25cm}
} 
\subfloat{
	\label{f0d}
	\includegraphics[width=\scaleFactor\textwidth]{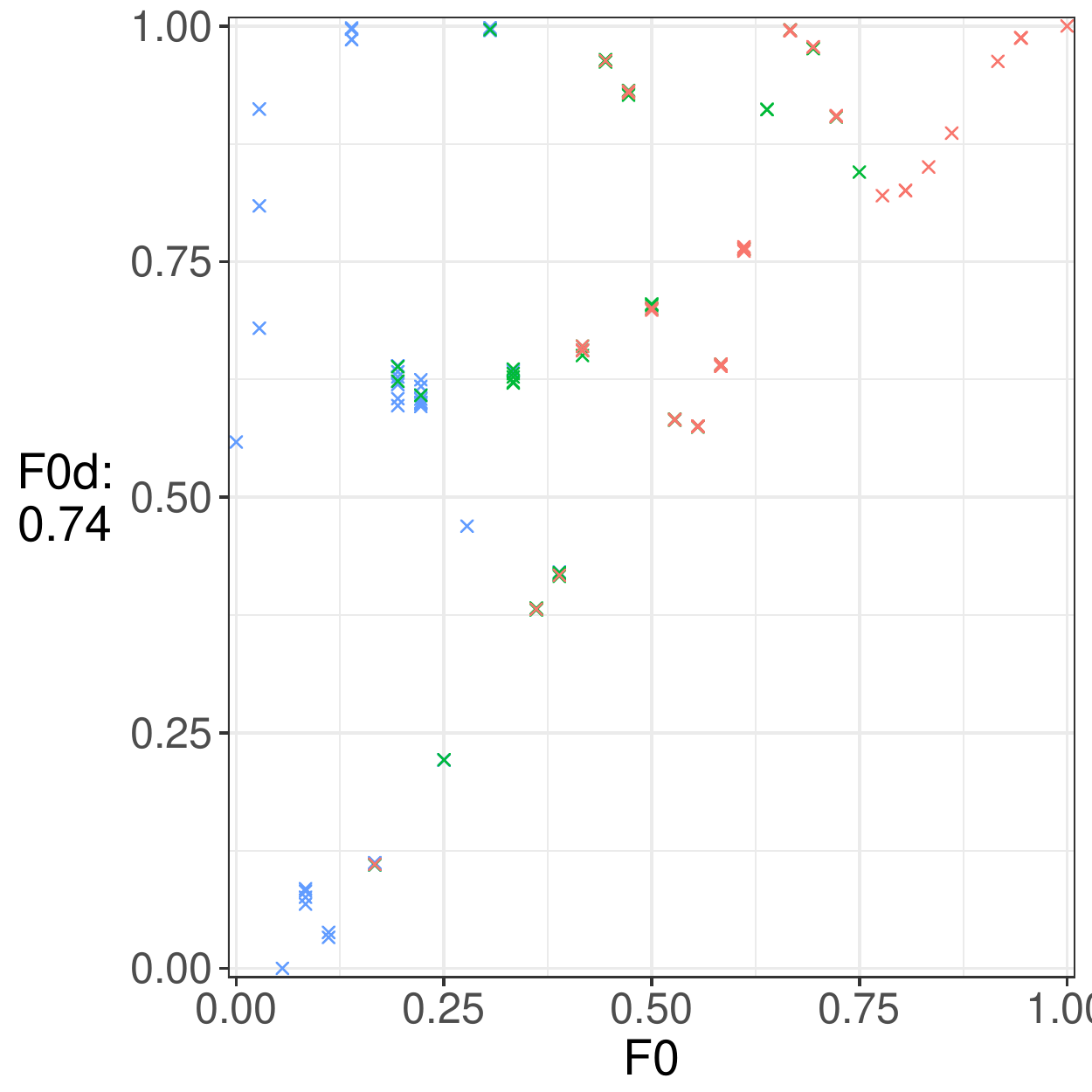}
\hspace{-.25cm}
} 
\subfloat{
	\label{f0e}
	\includegraphics[width=\scaleFactor\textwidth]{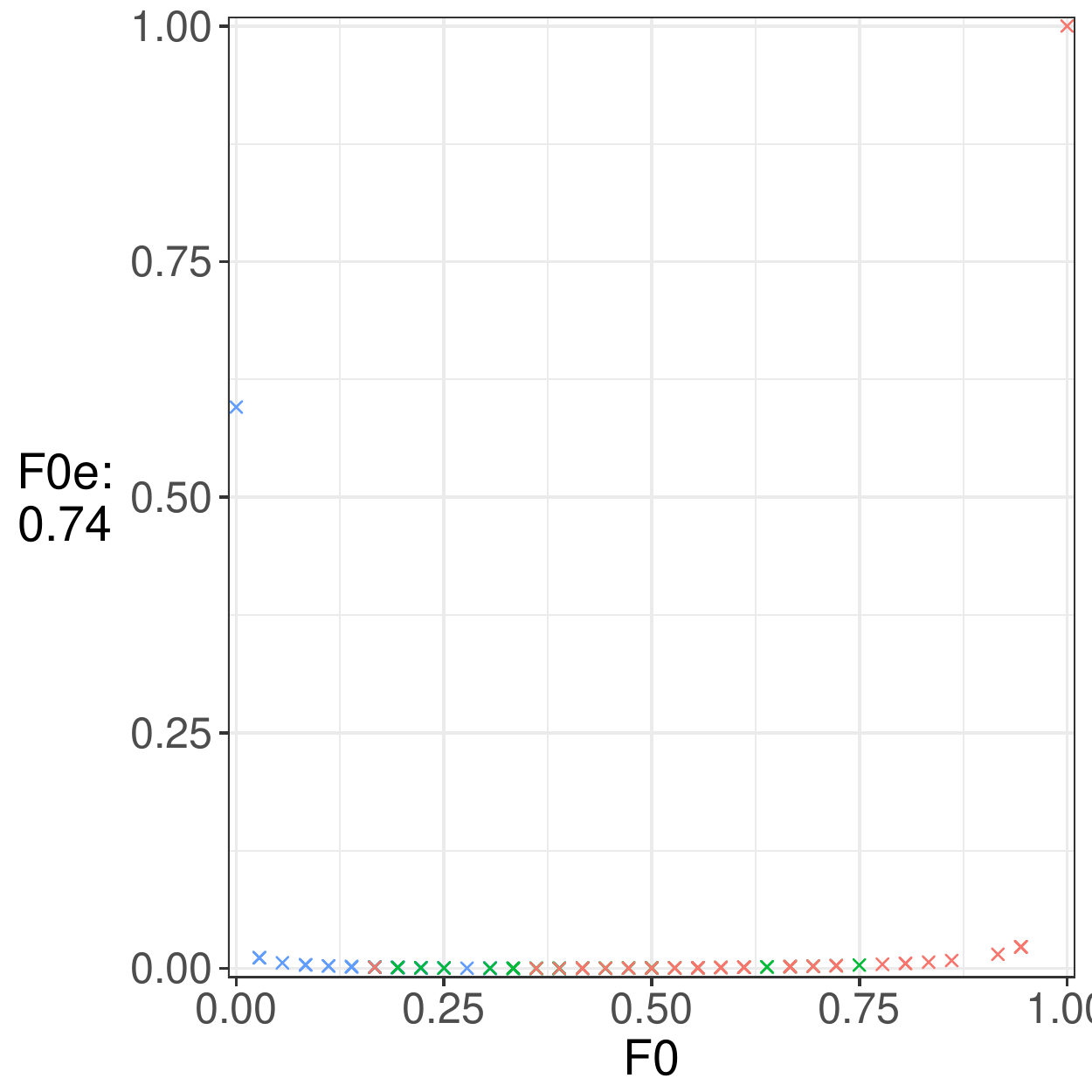}
}\vspace{-1em}

	\subfloat{
	\label{f1a}
	\includegraphics[width=\scaleFactor\textwidth]{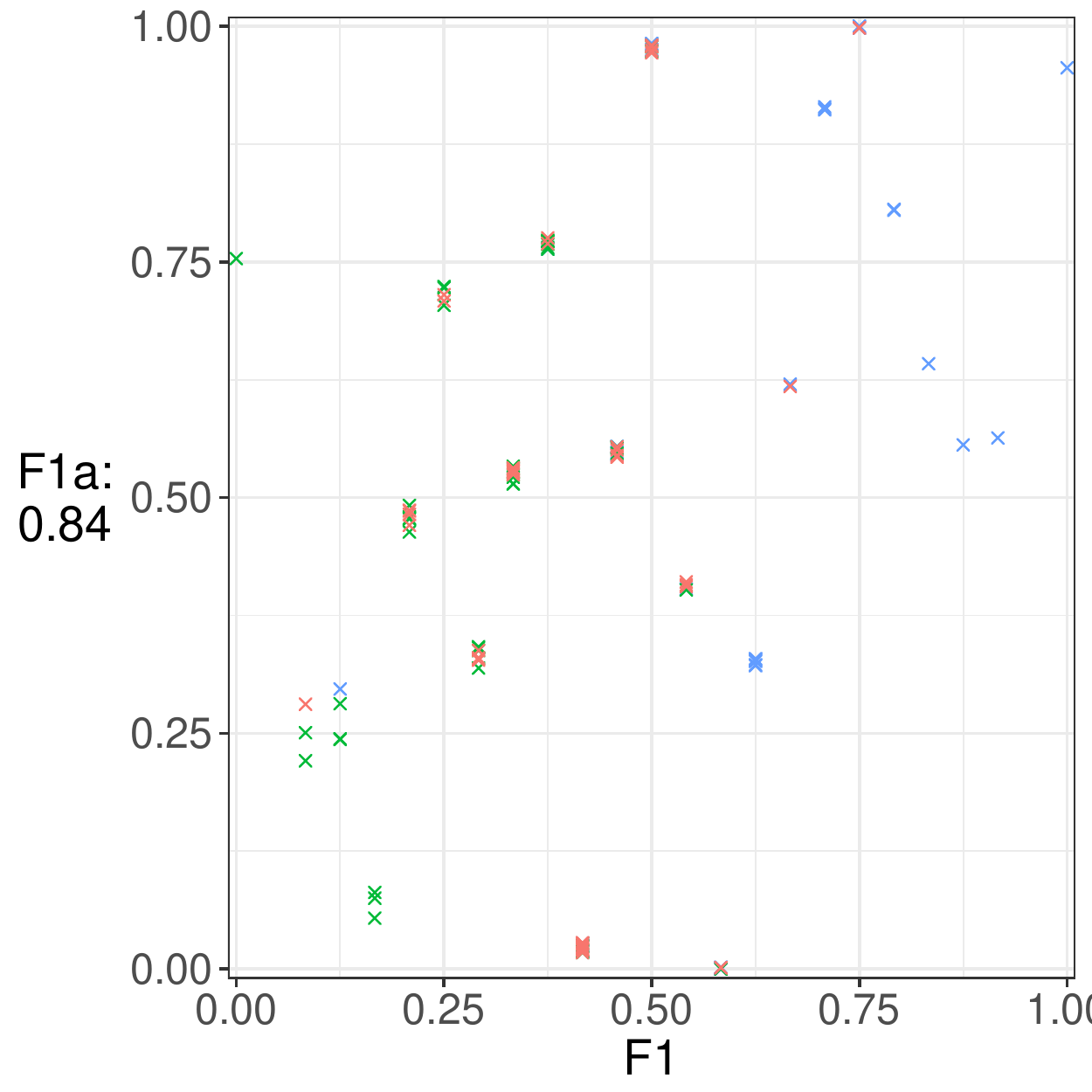}
\hspace{-.25cm}
} 
\subfloat{
	\label{f1b}
	\includegraphics[width=\scaleFactor\textwidth]{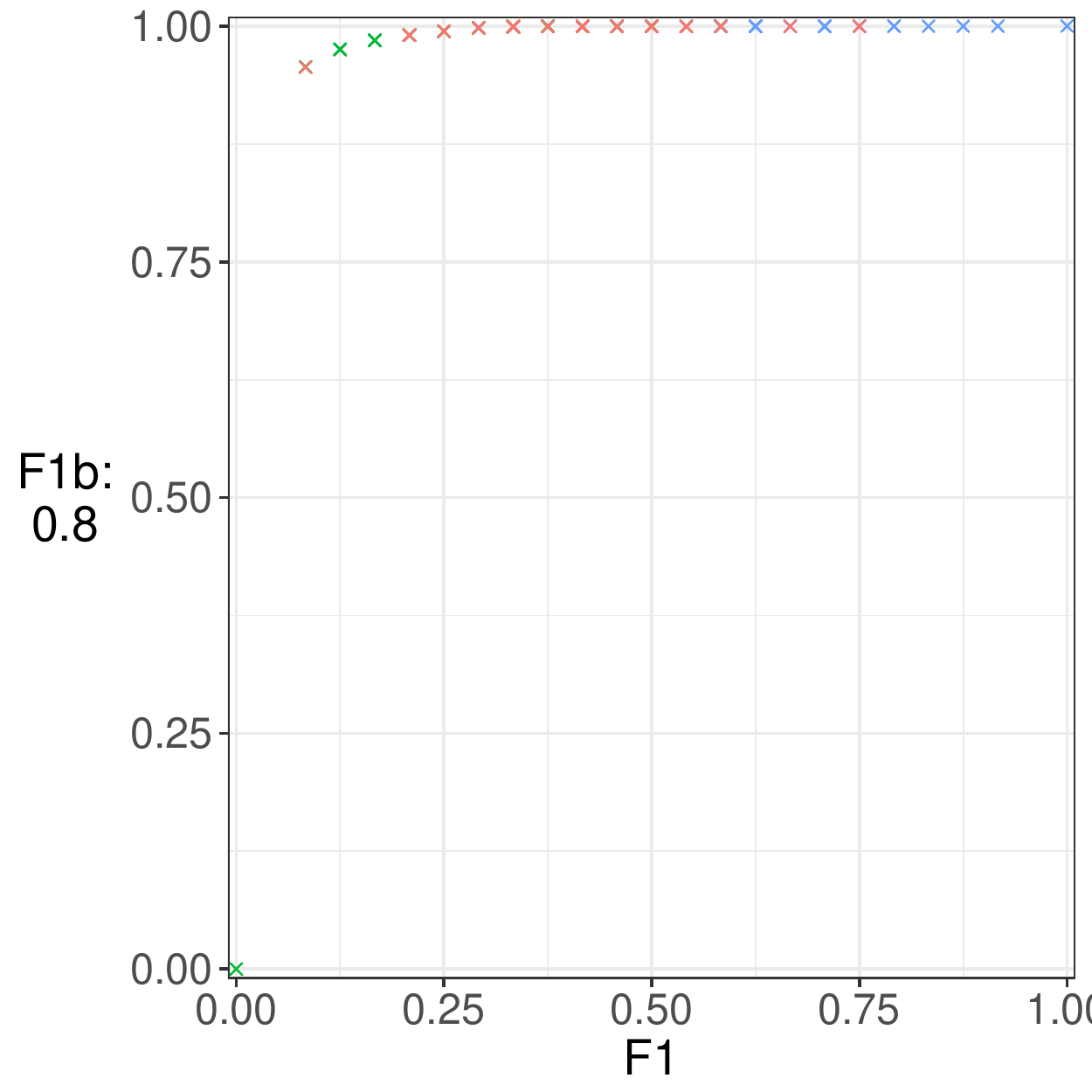}
\hspace{-.25cm}
} 
\subfloat{
	\label{f1c}
	\includegraphics[width=\scaleFactor\textwidth]{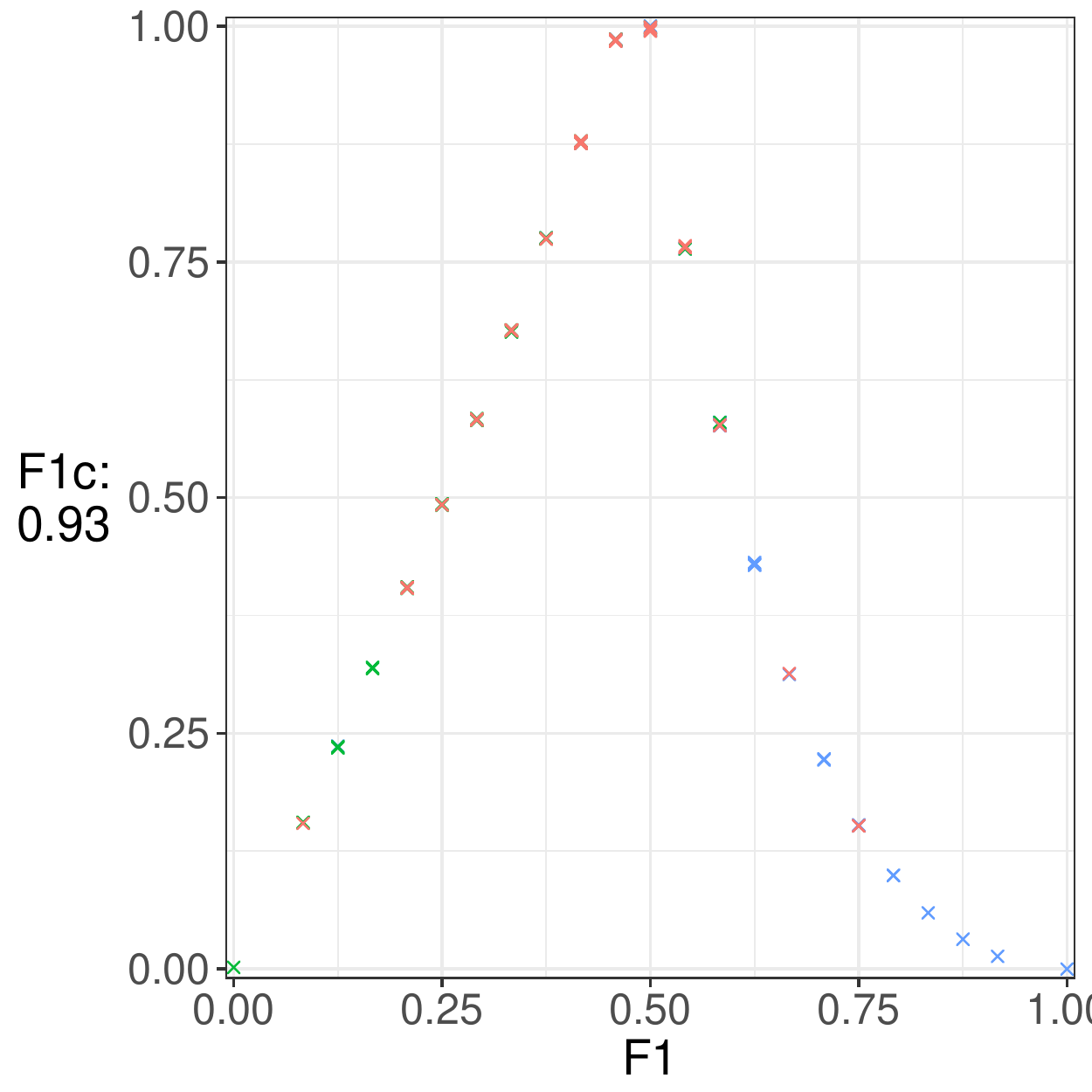}
\hspace{-.25cm}
} 
\subfloat{
	\label{f1d}
	\includegraphics[width=\scaleFactor\textwidth]{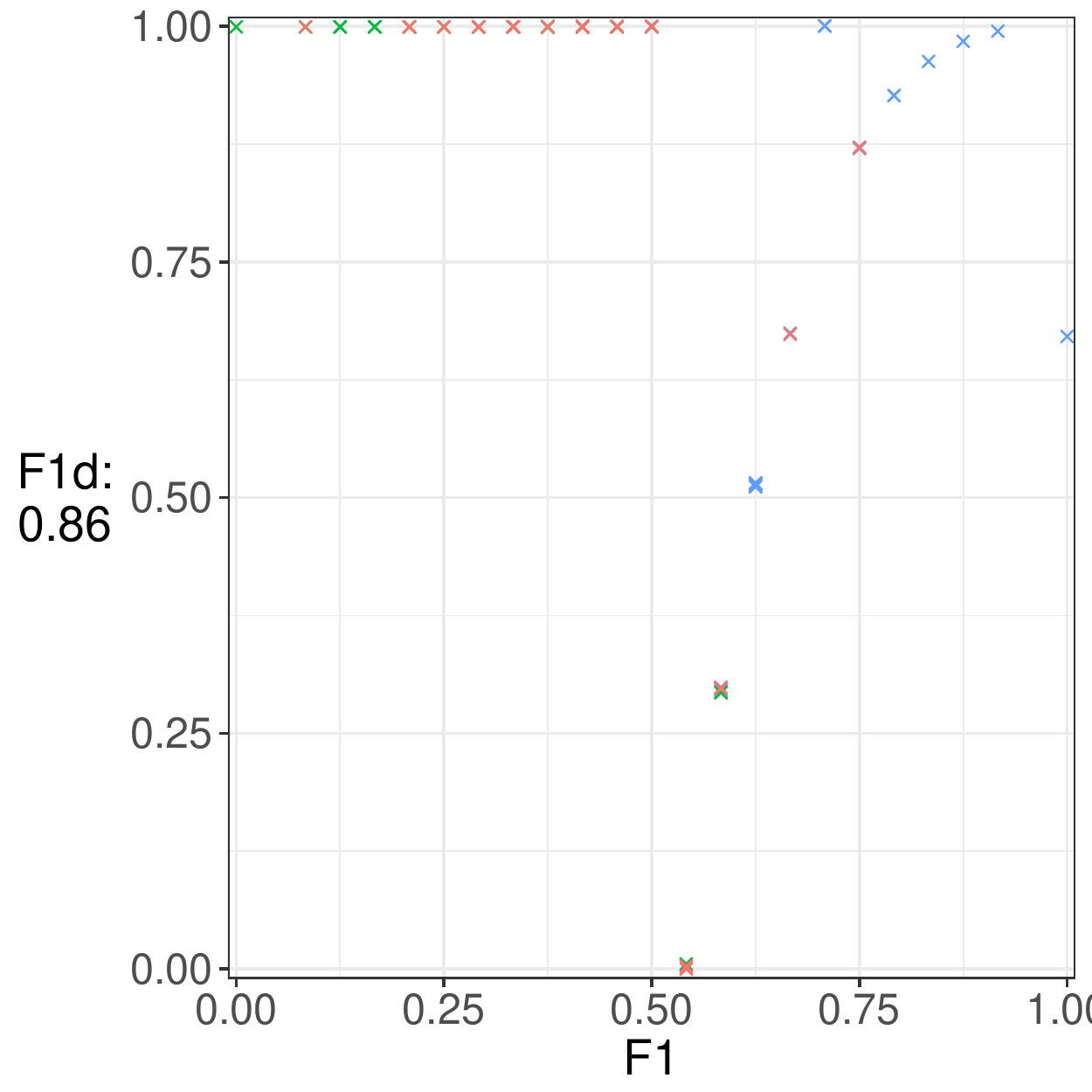}
\hspace{-.25cm}
} 
\subfloat{
	\label{f1e}
	\includegraphics[width=\scaleFactor\textwidth]{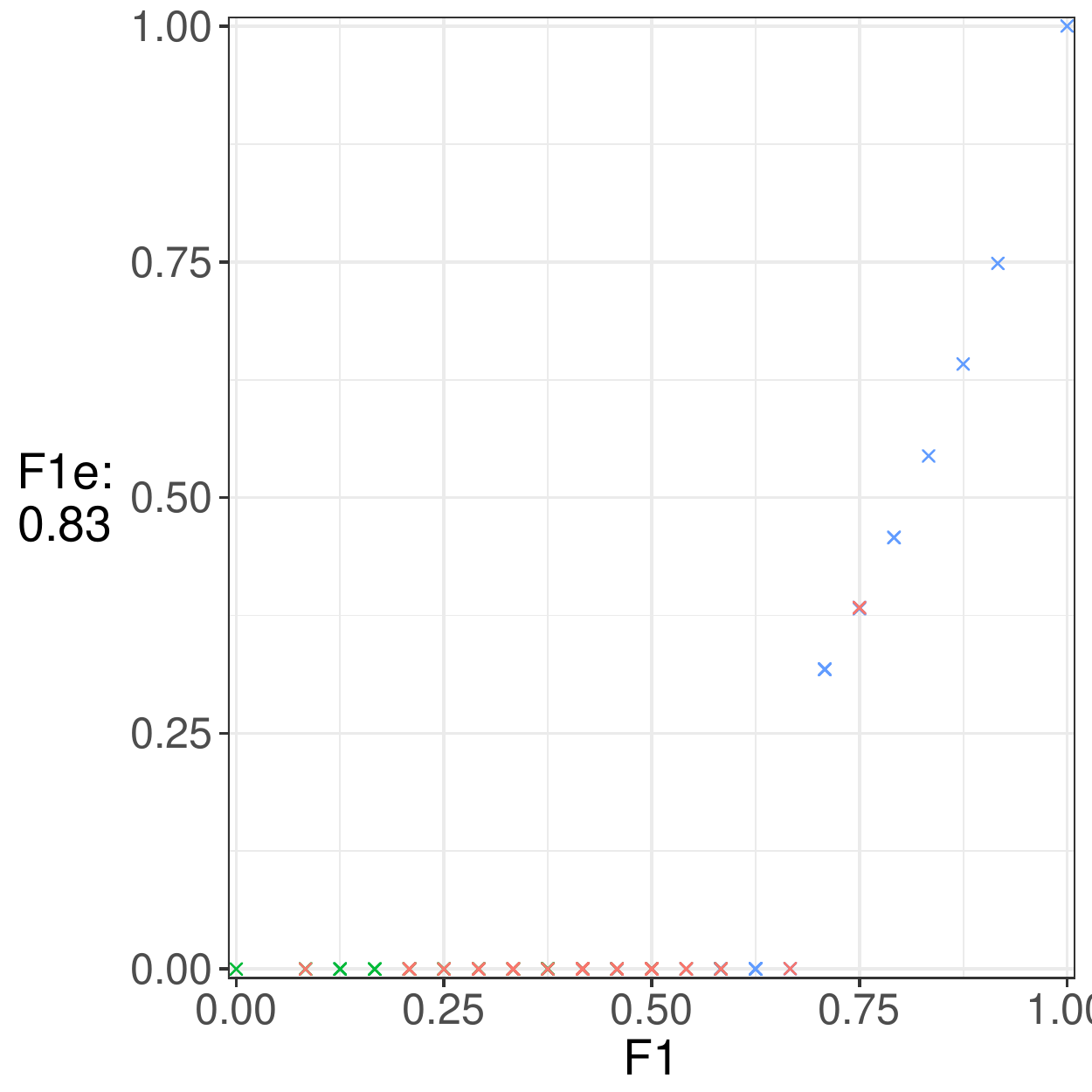}
}\vspace{-1em}

	\subfloat{
	\label{f2a}
	\includegraphics[width=\scaleFactor\textwidth]{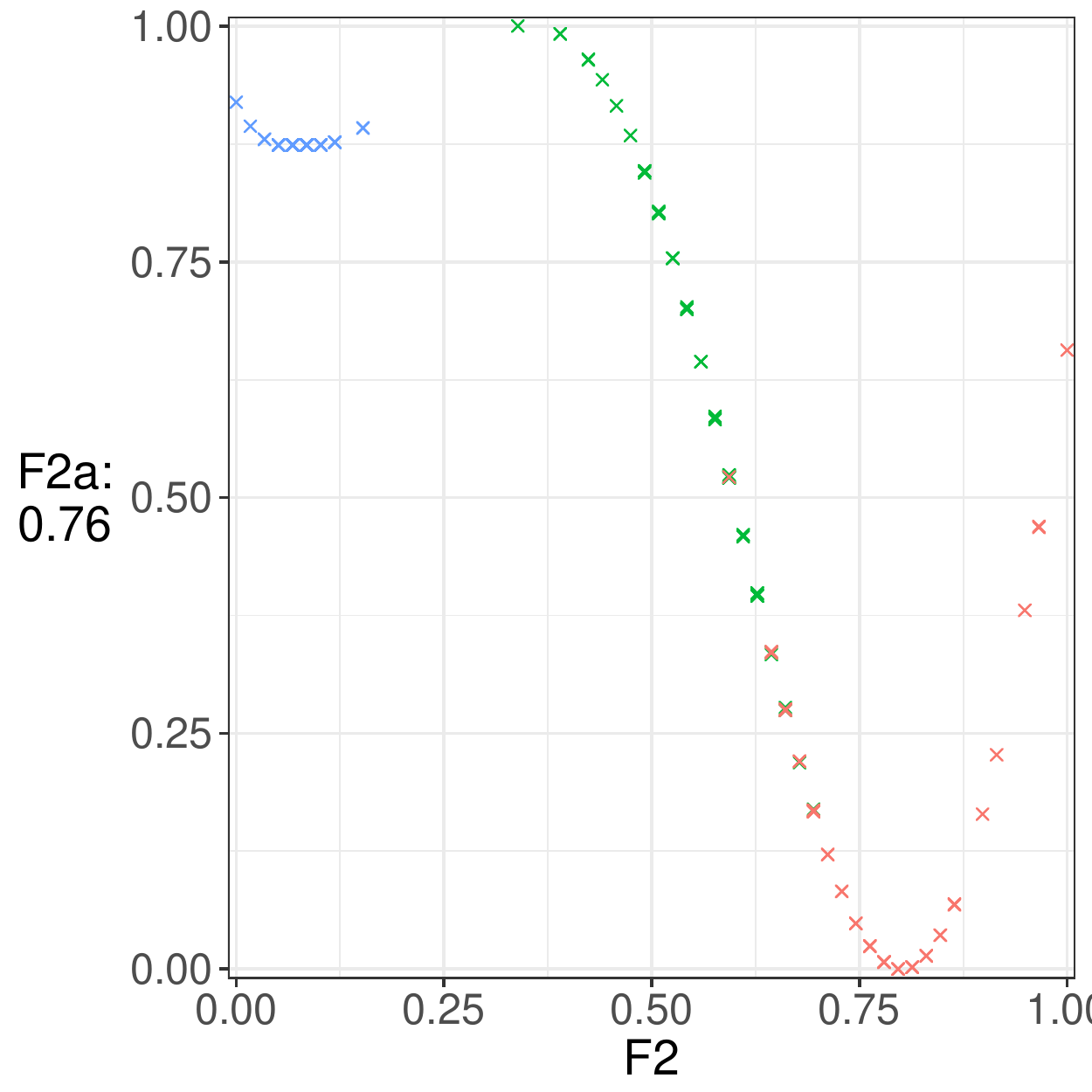}
\hspace{-.25cm}
} 
\subfloat{
	\label{f2b}
	\includegraphics[width=\scaleFactor\textwidth]{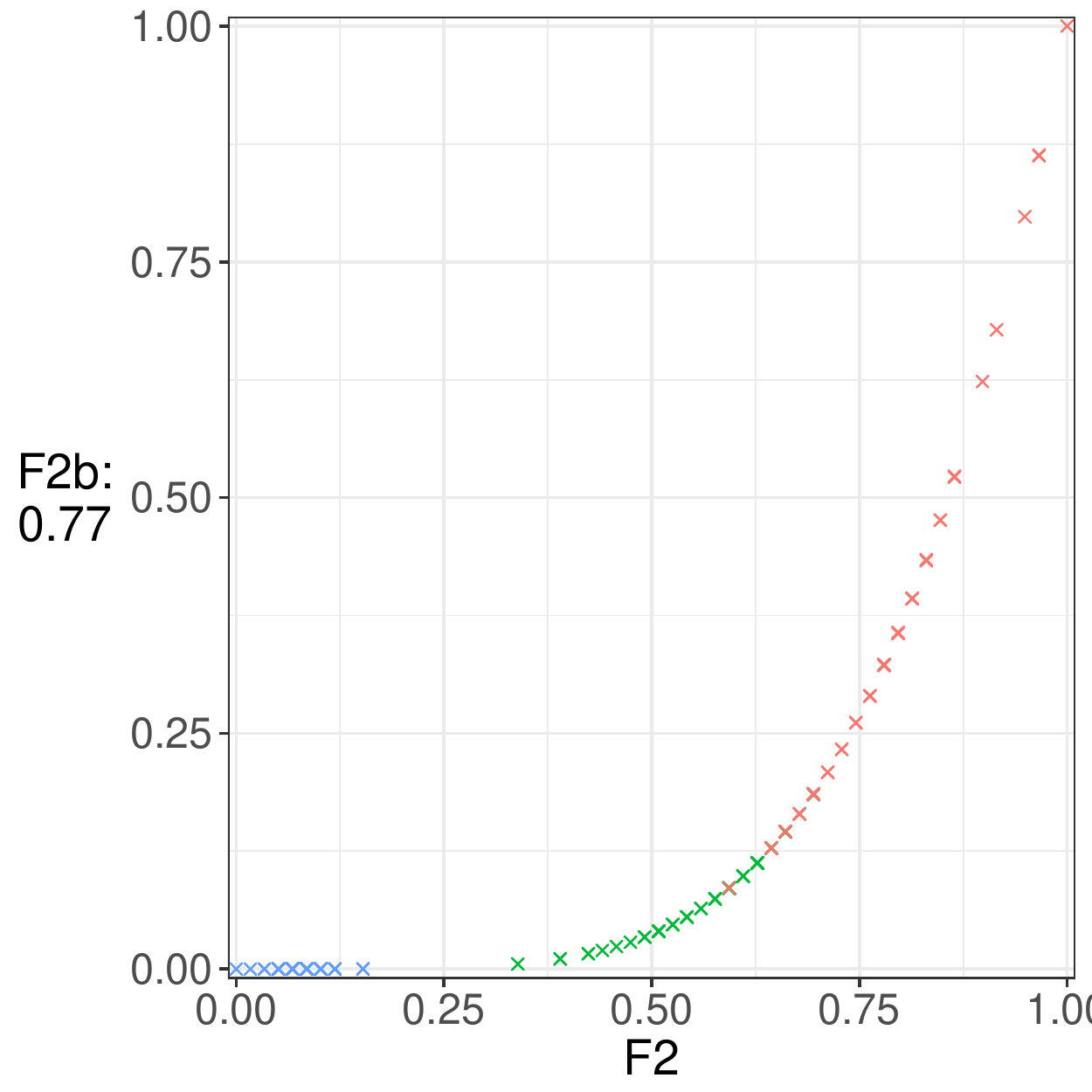}
\hspace{-.25cm}
} 
\subfloat{
	\label{f2c}
	\includegraphics[width=\scaleFactor\textwidth]{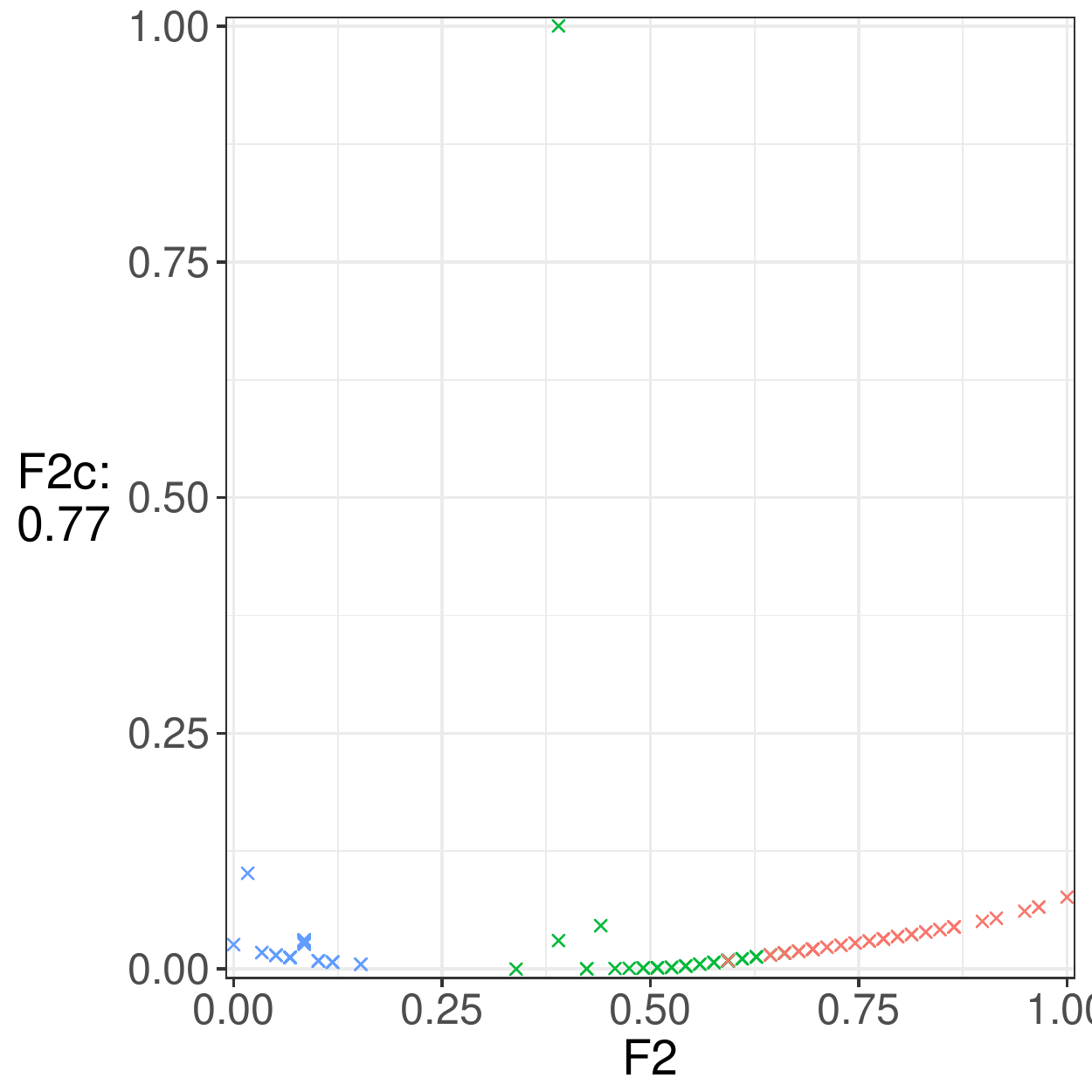}
\hspace{-.25cm}
} 
\subfloat{
	\label{f2d}
	\includegraphics[width=\scaleFactor\textwidth]{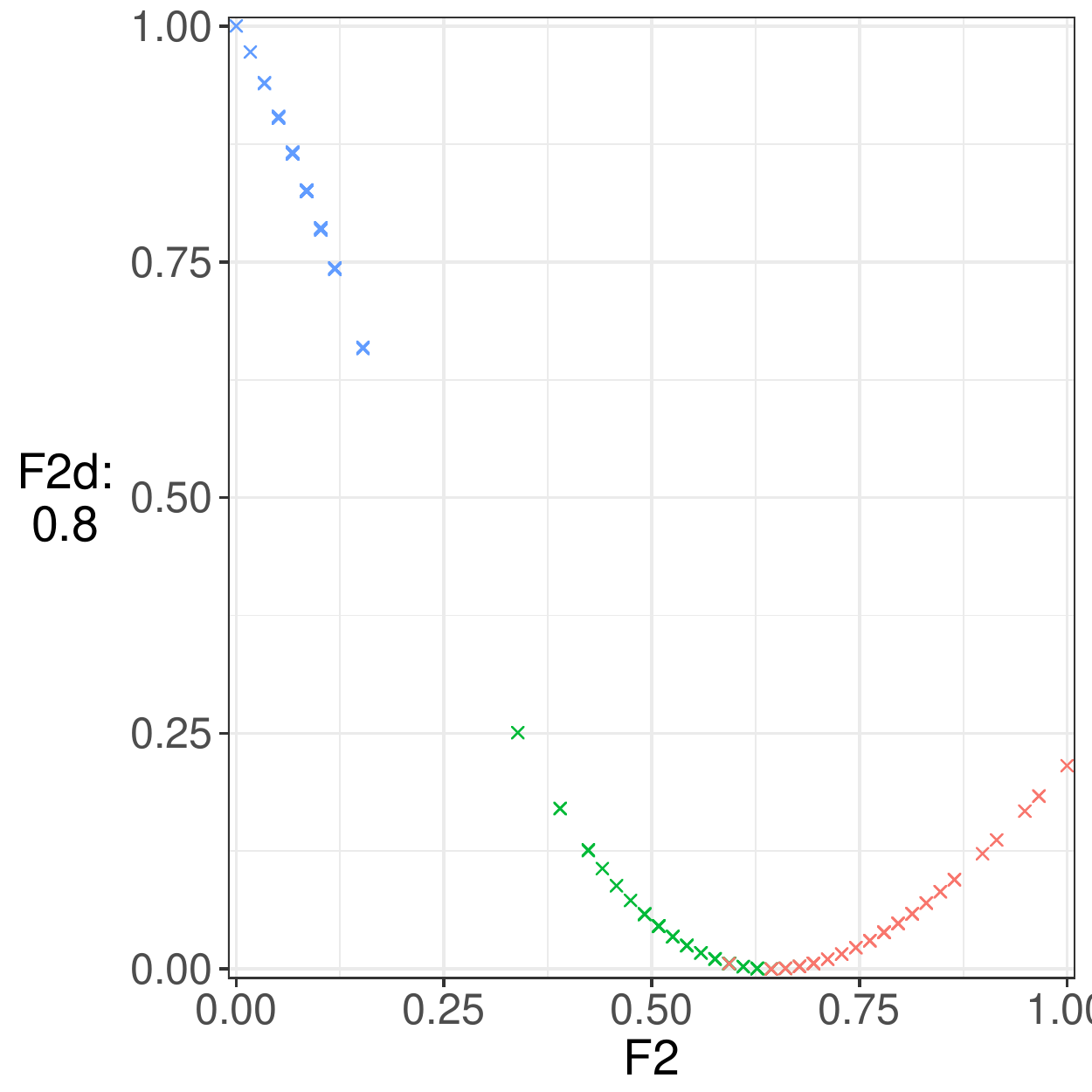}
\hspace{-.25cm}
} 
\subfloat{
	\label{f2e}
	\includegraphics[width=\scaleFactor\textwidth]{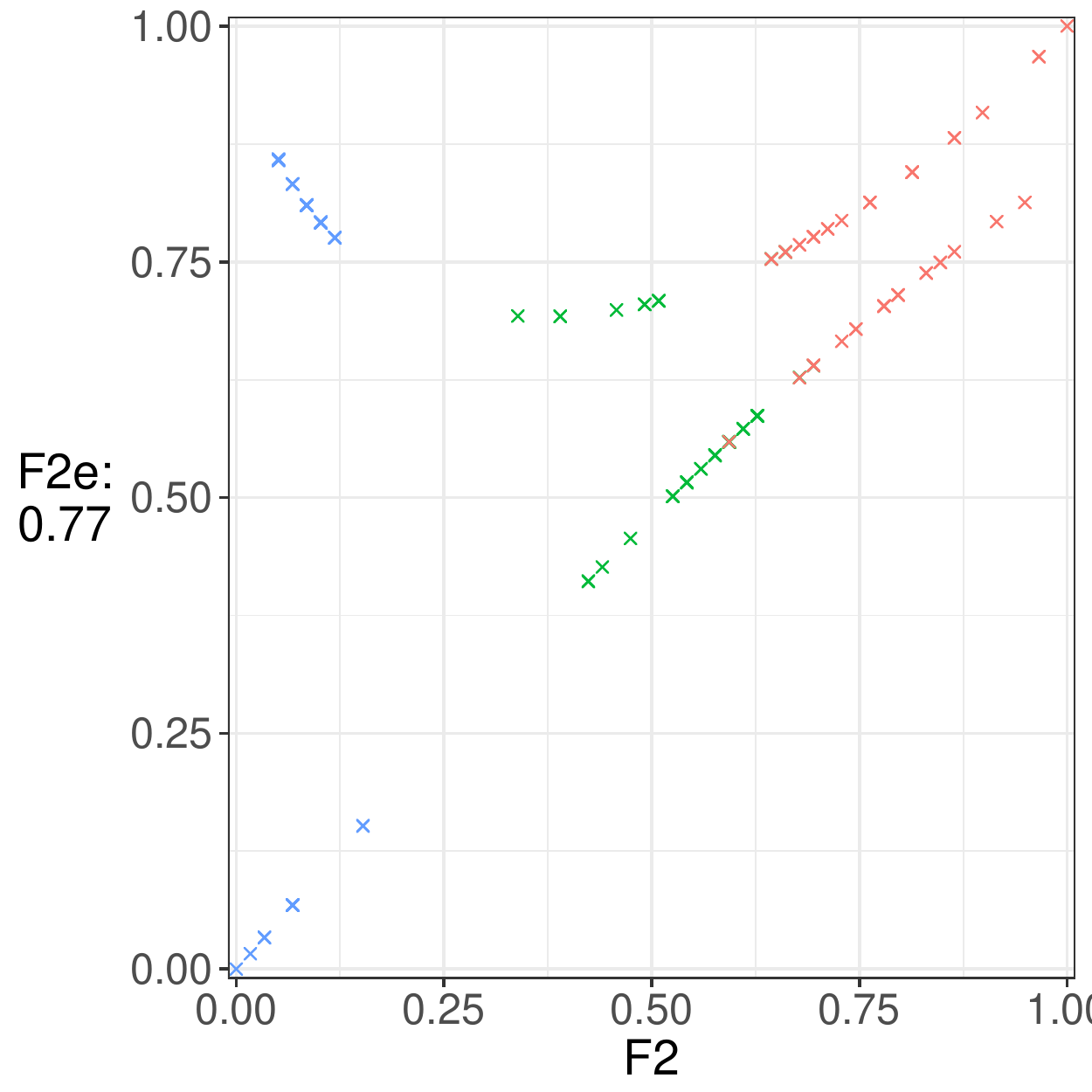}
}\vspace{-1em}

	\subfloat{
	\label{f3a}
	\includegraphics[width=\scaleFactor\textwidth]{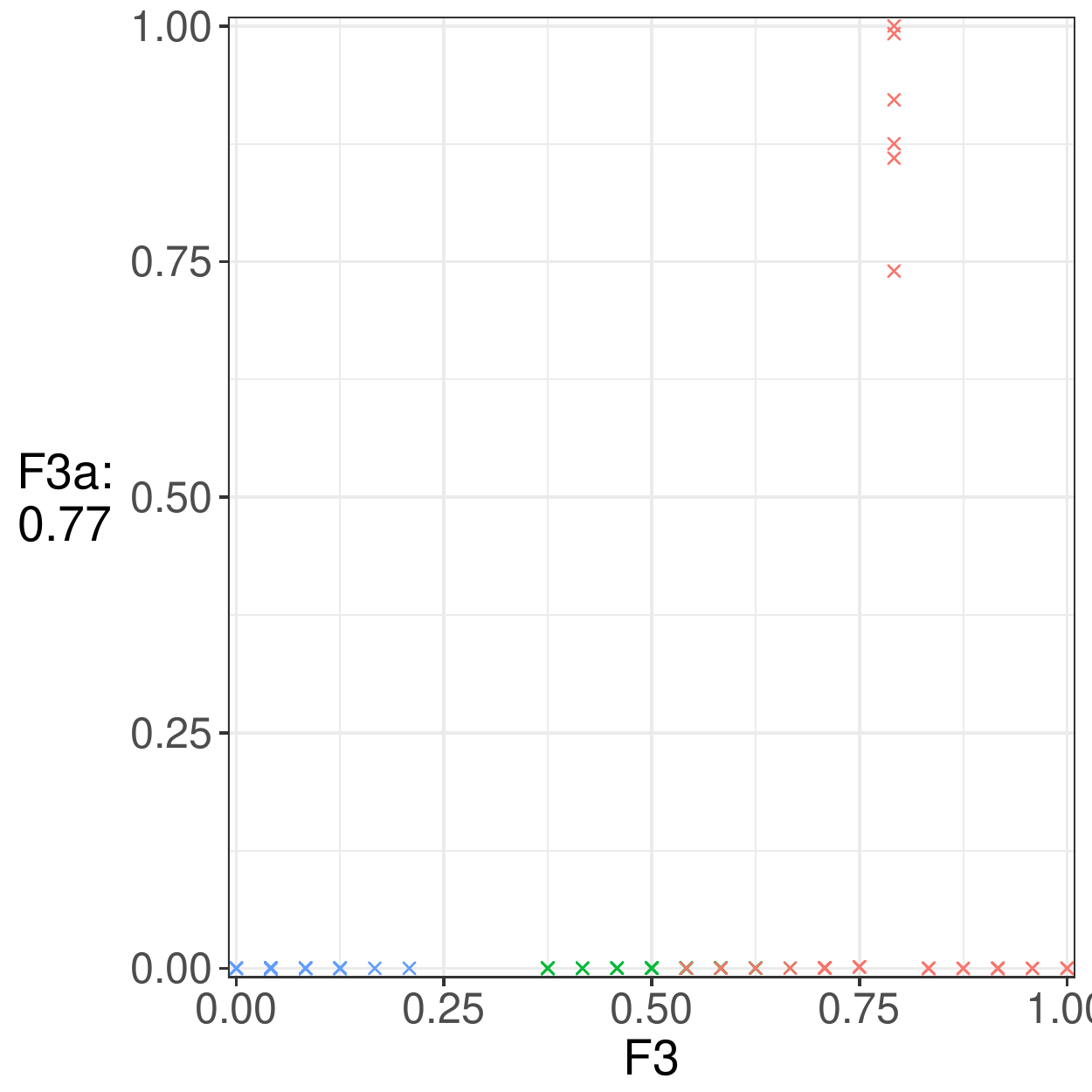}
\hspace{-.25cm}
} 
\subfloat{
	\label{f3b}
	\includegraphics[width=\scaleFactor\textwidth]{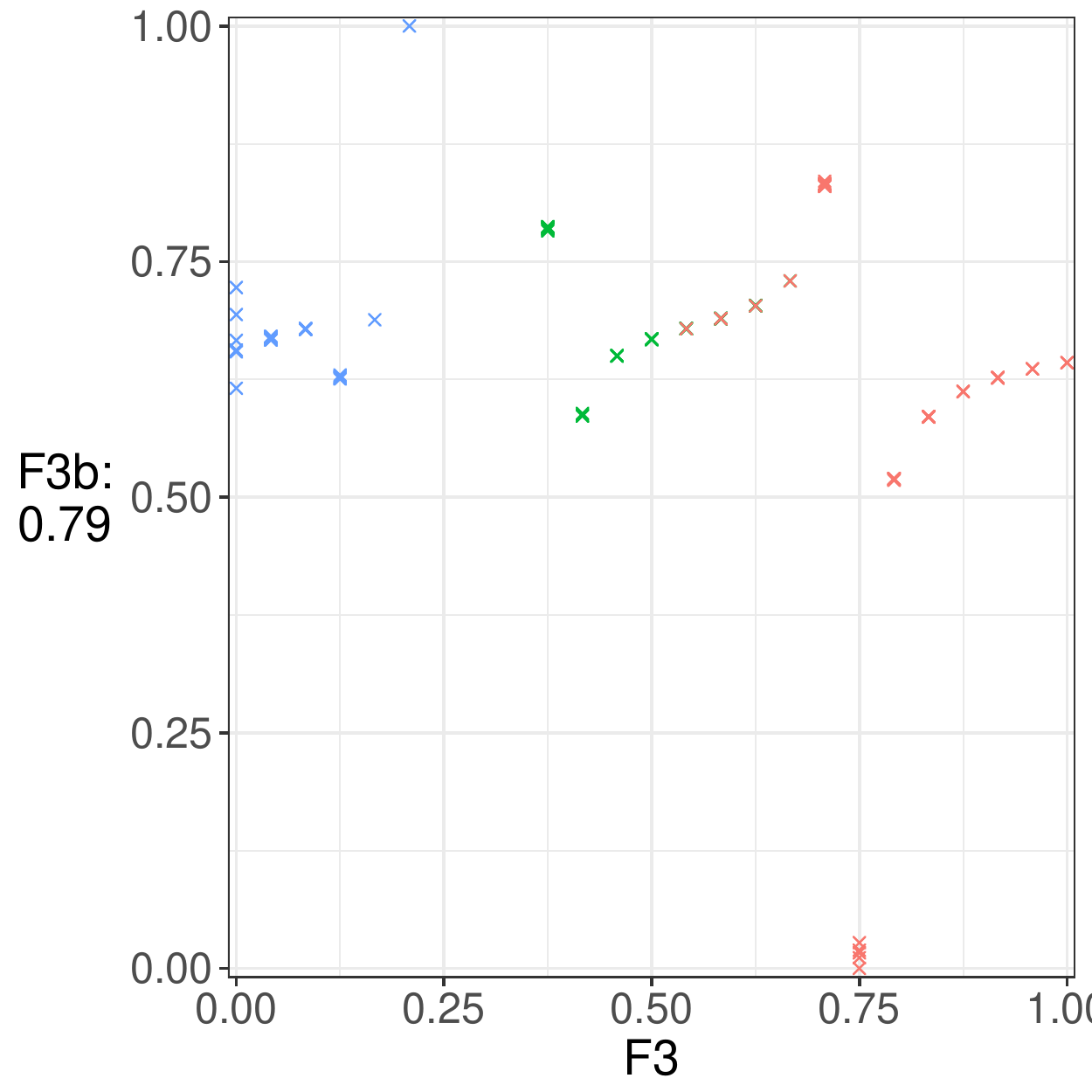}
\hspace{-.25cm}
} 
\subfloat{
	\label{f3c}
	\includegraphics[width=\scaleFactor\textwidth]{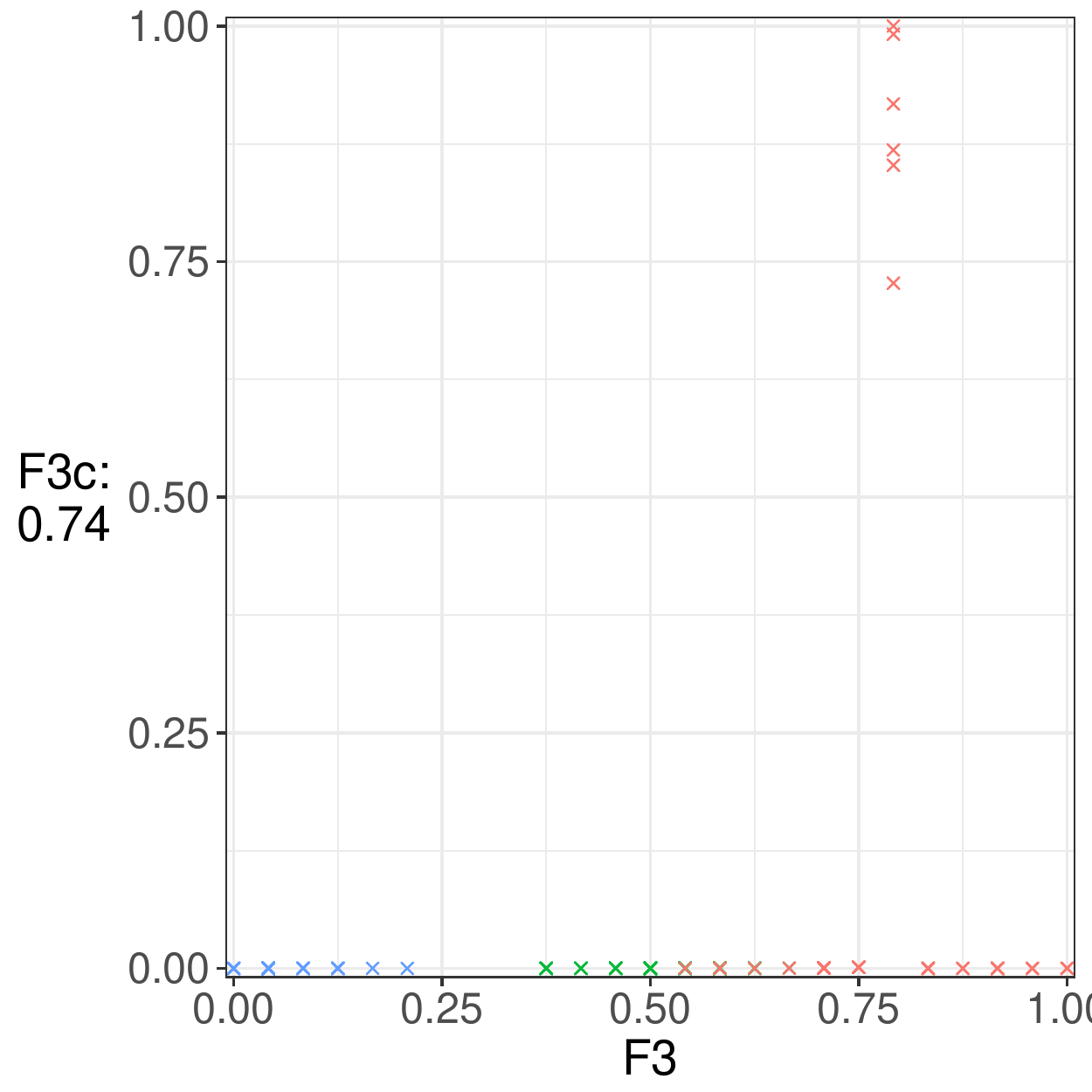}
	\hspace{-.25cm}
} 
\subfloat{
	\label{f3d}
	\includegraphics[width=\scaleFactor\textwidth]{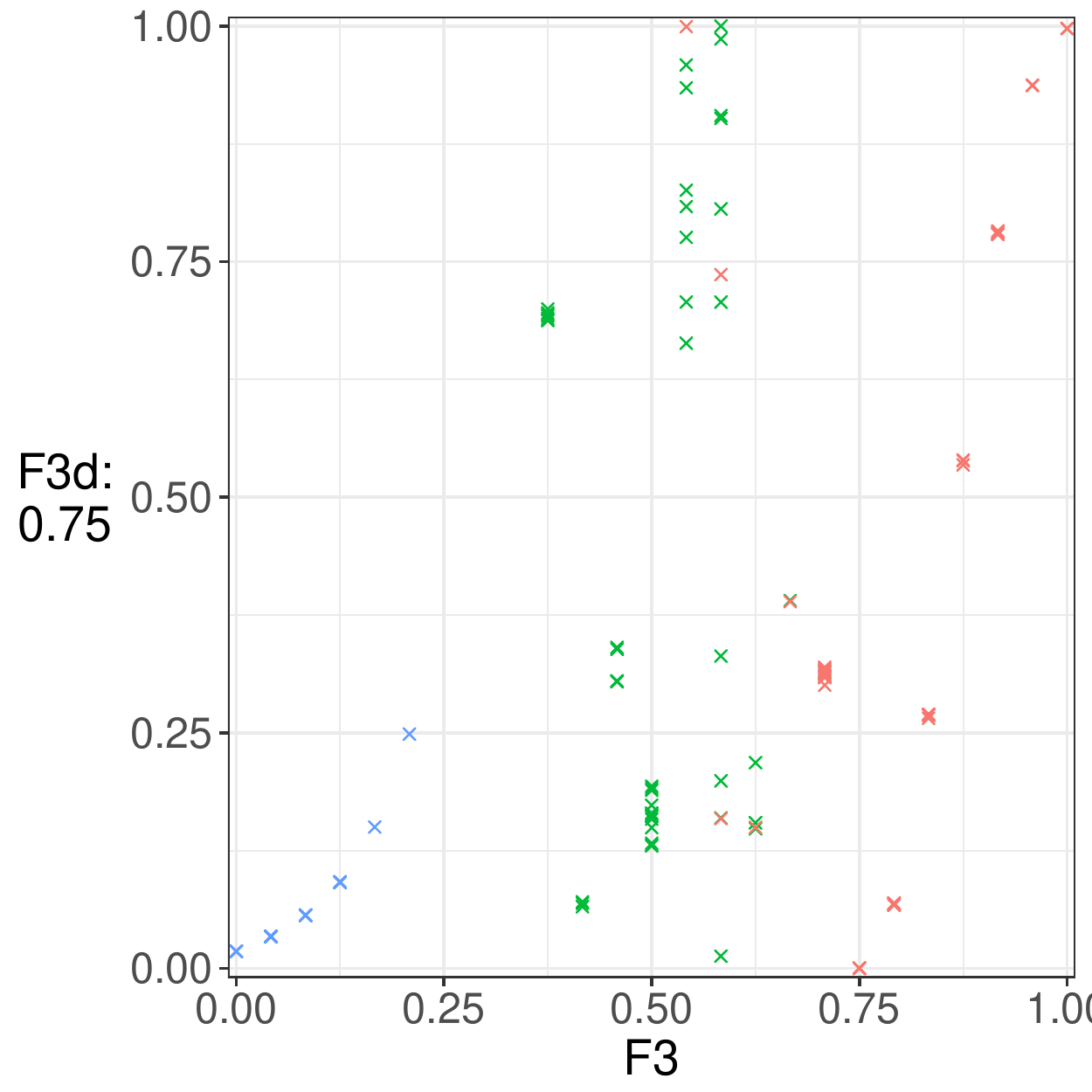}
\hspace{-.25cm}
} 
\subfloat{
	\label{f3e}
	\includegraphics[width=\scaleFactor\textwidth]{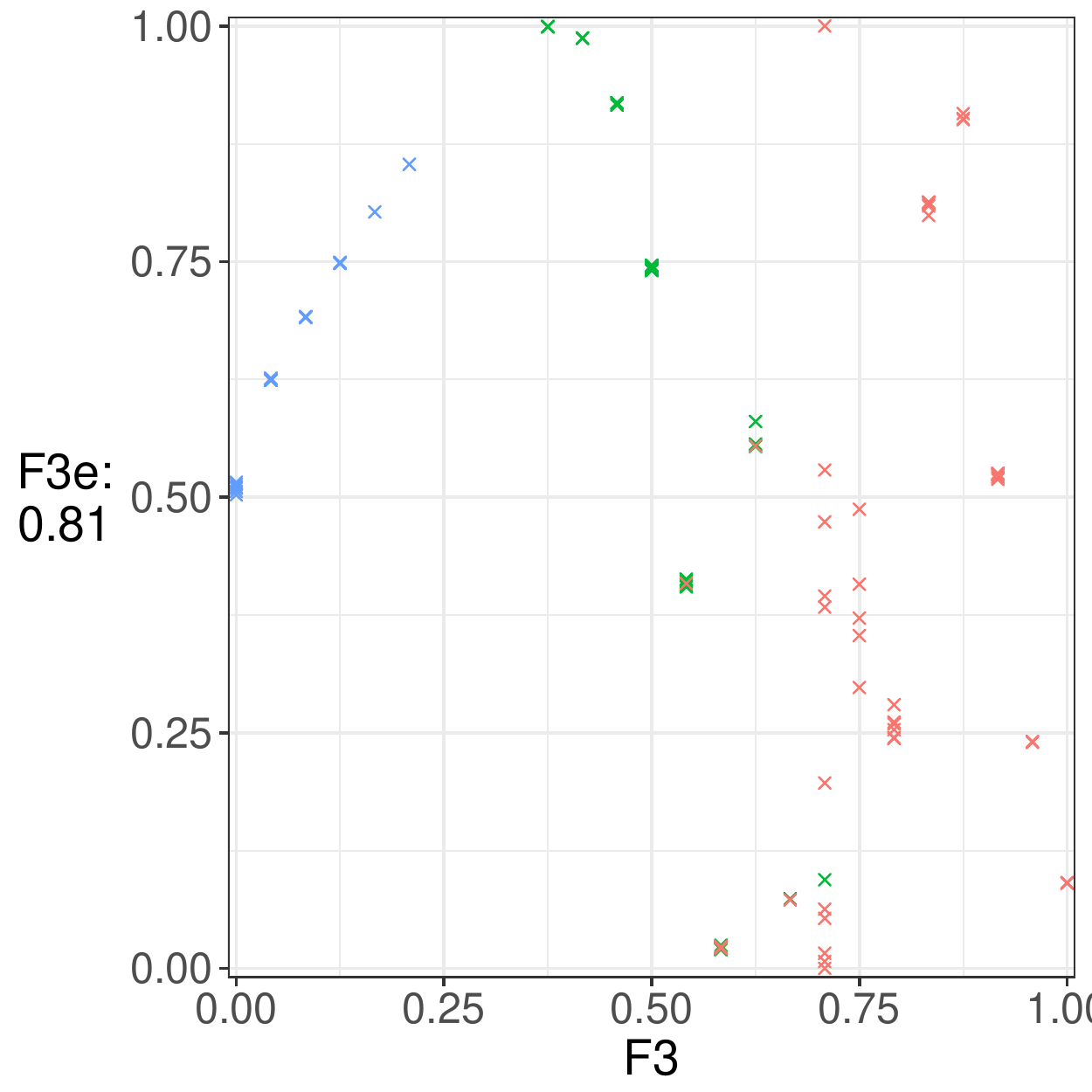}
}%\vspace{-1em}
	
	\hspace{2cm}\caption{Source Feature (x-axis) plotted against the five r.fs (y-axis) for each of F0 (1st row) to F3 (4th row) on Iris. The y-axis title is the name of each r.f and its source MI. Points are coloured red, green, or blue to indicate they belong to the setosa, versicolor, or virginica classes respectively. A small amount of jitter is added to each point to distinguish overlapping points.}
	\label{irisPlots}
%	\vspace{-2em}
\end{sidewaysfigure}

The most striking observation of these plots is that the functions produced by GPRFC are incredibly varied --- in fact, nearly every plot has a distinct appearance. The functions are also clearly complex, with no linear relationships apparent. A few functions are somewhat recognisable: for example, F2a (similar to a sine wave), F2b (a power curve), and F2d (a polynomial). The function evolved for F0a is similar in appearance to a sigmoid function, despite the sigmoid not being directly in the function set. Many of the remaining functions are more difficult to classify, as they either appear to have a number of different components (e.g.\  F0b, F3b), or have a majority of instances at a similar scale (e.g.\ F0e, F1b, F3a, F3c).

While generally each set of r.fs for a given source feature appear to be quite distinct, F3a and F3c appear to be very similar. This behaviour is counter-intuitive, as the fitness function directly penalises a r.f being similar to other r.fs for the same source feature. Indeed, F3a and F3c have a MI of 0.83 --- however, they each have very low MI (a maximum of 0.14) with the other r.fs (F3b/d/e), which means their average shared MI is still very low, at 0.34. The two trees corresponding to these two features are very similar (see Figure \ref{figure:f3af3c}). This issue may be alleviated by adapting the fitness function to consider the \textbf{worst-case}: that is, what is the highest value a given r.f shares with another r.f?

\begin{figure}[tb]
	
	\begin{verbatim}
	F3a = (pow (+ (exp (sqrt
	(cube X))) 
	(neg (log X))) (tan (exp (+ (log X) (mul X X)))))
	\end{verbatim}
	\begin{verbatim}
	F3c = (pow (+ (exp (sqrt 
	(sqrt (tan (square (square (max (sin (exp X)) (sin X)))))))) 
	(neg (log X))) (tan (exp (+ (log X) (mul X X)))))
	\end{verbatim}
	\caption{The example trees produced by GPRFC for F3a and F3c. The entire first and third lines are shared by both trees.}
	\label{figure:f3af3c}
	%	\vspace{-1em}
\end{figure}
%iris.data23FC.e3144231

%While the above results show the proposed method can generate good, difficult r.fs, and has potential as a means for generating benchmark datasets, it is unclear as to \textbf{why} the created r.fs are useful. To investigate this aspect, we will look at some examples of very good created r.fs in terms of their relationship with the source features.
%
%Figures \ref{fig:irisf2rfexample} and \ref{fig:irisf3rfexample} show an example of the generated r.fs for F2 and F3 respectively. The r.fs for F2 have a minSourceMI of 61\% and a maxSharedMI of 35\%, to give a fitness of 0.26. The r.fs for F3 have values of 63\%, 27\% and 0.36 respectively. As can be seen in the figures, the constructed r.fs are clearly quite complex, with a range of linear, polynomial, exponential and discontinuous functions used to map the source feature to the new r.fs.
%\begin{figure}[tp]
%	\centering
%	\includegraphics[width=0.99\linewidth]{/home/lensenandr/phd/reports/irisF2RFExample}
%	\caption{The distribution of the source feature F2 (top, histogram), and the 5 generated r.fs (bottom, each plotted as one coloured line).}
%	\label{fig:irisf2rfexample}
%\end{figure}
%\begin{figure}[tp]
%	\centering
%	\includegraphics[width=0.99\linewidth]{/home/lensenandr/phd/reports/irisF3RFExample}
%	\caption{The distribution of the source feature F3 (top, histogram), and the 5 generated r.fs (bottom, each plotted as one coloured line).}
%	\label{fig:irisf3rfexample}
%\end{figure}

\section{Conclusion}
\label{sec:conclusions}

This paper proposed the first approach to automatically evolving redundant features, using a Genetic Programming approach with a multi-tree representation, and a novel mutual information-based fitness function. The proposed GPRFC method was shown to generate high-quality and complex redundant features which are suitable for augmenting existing datasets for use in testing feature selection algorithms. We showed that good and interesting results could be achieved on both supervised and unsupervised problems. This paper represents the first piece of work in this area, but it already demonstrates the considerable potential of GP for this task. We hope that others in the GP community share our optimism, and we expect GP to ultimately be able to generate good benchmark data sets that can be used to test FS methods in data mining tasks such as classification, clustering and regression.

As GP has not been used for this sort of task previously, there is a number of different extensions that could be researched in the future. There is certainly scope for refining the fitness function further, in order to produce even more complex and distinct sets of r.fs. This work considered only one-to-one feature redundancies --- the source feature to each of the r.fs in turn. More difficult/complex feature redundancy relationships could be formed by using a multivariate mutual information approach, where a set of multiple source features are used to create a set of r.fs, i.e. many-to-many redundancies. The GP representation could also be refined further, by investigating more rigorously which function set is most suitable to produce good r.fs, and evaluating how the number of trees used is best determined.

\section*{Acknowledgements}
The authors would like to thank Tony Butler-Yeoman for his help in developing the initial ideas, and suggestions throughout the development of this work.

\bibliographystyle{splncs}
\bibliography{lensen.bib}

\begin{thebibliography}{10}

\bibitem{liu2012feature}
Liu, H., Motoda, H.:
\newblock Feature selection for knowledge discovery and data mining. Volume
  454.
\newblock Springer Science \& Business Media (2012)

\bibitem{tang2014feature}
Tang, J., Alelyani, S., Liu, H.:
\newblock Feature selection for classification: {A} review.
\newblock In: Data Classification: Algorithms and Applications.
\newblock (2014)  37--64

\bibitem{xue2015survey}
Xue, B., Zhang, M., Browne, W.N., Yao, X.:
\newblock A survey on evolutionary computation approaches to feature selection.
\newblock {IEEE} Trans. Evolutionary Computation \textbf{20}(4) (2016)
  606--626

\bibitem{espejo2010survey}
Espejo, P.G., Ventura, S., Herrera, F.:
\newblock A survey on the application of genetic programming to classification.
\newblock {IEEE} Trans. Systems, Man, and Cybernetics, Part {C} \textbf{40}(2)
  (2010)  121--144

\bibitem{jaynes1957information}
Jaynes, E.T.:
\newblock Information theory and statistical mechanics.
\newblock Physical review \textbf{106}(4) (1957)  620

\bibitem{kraskov2004estimating}
Kraskov, A., St{\"o}gbauer, H., Grassberger, P.:
\newblock Estimating mutual information.
\newblock Physical review E \textbf{69}(6) (2004)  066138

\bibitem{lizier2014jidt}
Lizier, J.T.:
\newblock {JIDT:} an information-theoretic toolkit for studying the dynamics of
  complex systems.
\newblock Front. Robotics and {AI} \textbf{2014} (2014)

\bibitem{tran2016genetic}
Tran, B., Xue, B., Zhang, M.:
\newblock Genetic programming for feature construction and selection in
  classification on high-dimensional data.
\newblock Memetic Computing \textbf{8}(1) (2016)  3--15

\bibitem{lensen2017gpgc}
Lensen, A., Xue, B., Zhang, M.:
\newblock {GPGC:} genetic programming for automatic clustering using a flexible
  non-hyper-spherical graph-based approach.
\newblock In: Proceedings of the Genetic and Evolutionary Computation
  Conference, {GECCO}., {ACM} (2017)  449--456

\bibitem{muni2006genetic}
Muni, D.P., Pal, N.R., Das, J.:
\newblock Genetic programming for simultaneous feature selection and classifier
  design.
\newblock {IEEE} Trans. Systems, Man, and Cybernetics, Part {B} \textbf{36}(1)
  (2006)  106--117

\bibitem{Ahmed2014Multiple}
Ahmed, S., Zhang, M., Peng, L., Xue, B.:
\newblock Multiple feature construction for effective biomarker identification
  and classification using genetic programming.
\newblock In: Proceedings of the Genetic and Evolutionary Computation
  Conference, {GECCO} '14, Vancouver, BC, Canada., {ACM} (2014)  249--256

\bibitem{YunZhang2004Multiple}
Zhang, Y., Zhang, M.:
\newblock A multiple-output program tree structure in genetic programming.
\newblock Technical report, Victoria University of Wellington, New Zealand
  (2004)

\bibitem{Lin2005Evolutionary}
Lin, Y., Bhanu, B.:
\newblock Evolutionary feature synthesis for object recognition.
\newblock {IEEE} Trans. Systems, Man, and Cybernetics, Part {C} \textbf{35}(2)
  (2005)  156--171

\bibitem{neshatian2012filter}
Neshatian, K., Zhang, M., Andreae, P.:
\newblock A filter approach to multiple feature construction for symbolic
  learning classifiers using genetic programming.
\newblock {IEEE} Trans. Evolutionary Computation \textbf{16}(5) (2012)
  645--661

\bibitem{uci}
Lichman, M.:
\newblock {UCI} machine learning repository (2013)

\bibitem{handl2007evolutionary}
Handl, J., Knowles, J.D.:
\newblock An evolutionary approach to multiobjective clustering.
\newblock {IEEE} Trans. Evolutionary Computation \textbf{11}(1) (2007)  56--76

\bibitem{hall2009weka}
Hall, M.A., Frank, E., Holmes, G., Pfahringer, B., Reutemann, P., Witten, I.H.:
\newblock The {WEKA} data mining software: an update.
\newblock {SIGKDD} Explorations \textbf{11}(1) (2009)  10--18

\bibitem{pudil1994floating}
Pudil, P., Novovicov{\'{a}}, J., Kittler, J.:
\newblock Floating search methods in feature selection.
\newblock Pattern Recognition Letters \textbf{15}(10) (1994)  1119--1125

\end{thebibliography}

\end{document}